\documentclass[final]{elsarticle}
\usepackage{tikz,graphics,color,epsf,float,epsf,caption,subcaption}
\usepackage{amsmath,amssymb}
\usepackage{amsmath}
\usepackage{bm}
\usepackage{amssymb}
\usepackage{multirow}
\usepackage{enumitem}
\usepackage{natbib}
\usepackage{url}
\usepackage{xcolor}
\usepackage{soul}
\journal{arXiv}
\biboptions{authoryear}

\usepackage{txfonts}
\usepackage[T1]{fontenc}

\begin{document}

\begin{frontmatter}

\title{SDCT-AuxNet$^{\theta}$: DCT Augmented Stain Deconvolutional CNN  with Auxiliary Classifier for Cancer Diagnosis}

\author[mymainaddress]{Shiv Gehlot}

\author[mymainaddress]{Anubha Gupta\corref{mycorrespondingauthor}}
\cortext[mycorrespondingauthor]{Corresponding author}
\ead{anubha@iiitd.ac.in}

\author[mysecondaddress]{Ritu Gupta\corref{mycorrespondingauthor}}
\ead{drritugupta@gmail.com }

\address[mymainaddress]{SBILab, Department of ECE, IIIT-Delhi, New Delhi, 110020, India}
\address[mysecondaddress]{Laboratory Oncology Unit, Dr. B.R.A.IRCH, AIIMS, New Delhi 110029, India}

\begin{abstract}
Acute lymphoblastic leukemia (ALL) is a pervasive pediatric white blood cell
cancer across the globe. With the popularity of convolutional neural networks
(CNNs), computer-aided diagnosis of cancer has attracted considerable attention. Such tools are easily deployable and are cost-effective. Hence, these can
enable extensive coverage of cancer diagnostic facilities. However, the development
of such a tool for ALL cancer was challenging so far due to the non-availability of
a large training dataset. The visual similarity between the malignant and normal
cells adds to the complexity of the problem. This paper discusses the recent
release of a large dataset and presents a novel deep learning architecture for the
classification of cell images of ALL cancer. The proposed architecture, namely,
\textit{SDCT-AuxNet}$^{\theta}$ is a 2-module framework that utilizes a compact CNN as the main classifier in one
module and a Kernel SVM as the auxiliary classifier in the other one. While CNN classifier uses
 features through bilinear-pooling, spectral-averaged features are used by the auxiliary  classifier.
Further, this CNN is trained on the stain deconvolved quantity images in the
optical density domain instead of the conventional RGB images. A novel test
strategy is proposed that exploits both the classifiers for decision making using the confidence scores of their predicted class labels. Elaborate experiments
have been carried out on our recently released public dataset of 15114 images of
ALL cancer and healthy cells to establish the validity of the proposed methodology that is also robust to subject-level variability. A weighted F1 score of 94.8$\%$
is obtained that is best so far on this challenging dataset.
\end{abstract}


\end{frontmatter}


\section{Introduction}
\label{sec:intro}
Cancer is a condition of unrestricted cell growth in the human body and is among the deadliest diseases of the world. In 2018, cancer claimed 9.6 million lives globally, $70\%$ of which are from low and middle-income countries (LMIC) \citep{ref1}. Of the different types of cancer, acute lymphoblastic leukemia (ALL)  is pervasive childhood cancer that occurs due to the excessive generation of immature white blood cell (WBC) blasts in the bone marrow.   Although prompt diagnosis is a critical factor in cancer survival, LMIC suffer on this account due to the lack of costly infrastructure, trained human resources, and diagnostic tools at the required scale. Hence, these countries experience higher fatality rates. For example, $84\%$ of cases of ALL are reported from LMIC \citep{ref2}.

Preliminary diagnostic tests of ALL are based on the count and appearance of blood cells. For example, \textit{complete blood count} (CBC) makes a diagnosis based on cell count, while \textit{peripheral blood smear} test examines the appearance of cells in the blood. Bone marrow based tests such as \textit{bone marrow biopsy} and \textit{bone marrow aspiration} are crucial in ALL diagnosis and analyze bone marrow to evaluate the symptoms of leukemia. Higher accuracy tests for the identification of leukemia include \textit{cytochemistry}, \textit{immunohistochemistry}, and \textit{flow cytometry} that are based on the reaction of staining chemicals with the proteins of blood cells. Imaging tests such as X-rays, computed tomography (CT) scan, magnetic resonance imaging (MRI) scan, and ultrasound are not useful for ALL detection, but they help with the analysis of the overall impact of cancer on the body \citep{ref3}.

All these methods are time-intensive, require costly medical equipment, and trained medical professionals. Owing to these requirements, these tools cannot be deployed easily at a large scale, especially in rural areas. Computer-aided tools based on microscopic image analysis can overcome these limitations because these can be fully automated and do not require highly trained medical professionals to run the tests.

Several machine learning-based computer-aided ALL diagnosis methods have been presented over the last few years \citep{Mohapatra2011,Monica2012,Joshi2013,ref9,Mohapatra2014,Chatap2014,SiewChin2015,Carolina2015,Vincent2015,PATEL2015,ref10,ref11,MoradiAmin2016,Singhal2016,ref58,RAWAT2017,Mishra2017,Karthikeyan2017,MISHRA2019}.  All these methods utilize a predefined set of features based on the structure of the nucleus or cytoplasm of the cells to train classifiers such as na\"{i}ve Bayes, decision tree, support vector machine (SVM), random forest, and the ensemble of classifiers for the diagnosis of ALL. A significant drawback of these methods is the use of tiny datasets (consisting of 19 to 267 images only) for building the classifiers. A classifier trained on a small dataset is prone to overfitting and hence, may not perform satisfactorily on the prospective subjects' samples. Also, the performance reported on a small test set may be biased.
Another limitation of these methods stems from the utilization of predefined/hand-crafted features that may not always be the optimized set of features. All these factors can limit the performance of the classifier on deployment in a real-life setting. These limitations can be overcome by using a larger dataset for building a classifier and by utilizing an automated mechanism to extract the features. 

Convolutional neural network (CNN) is a deep learning architecture consisting of two-dimensional convolutional filters that can check the limitation of predefined features by extracting task-specific features from the raw data through the minimization of a targeted loss function. In medical applications, CNNs are frequently used for classification \citep{ref22}, localization \citep{ref23}, detection \citep{ref24}, segmentation \citep{ref25}, and registration \citep{ref26}. In the recent past, numerous works have utilized CNNs for cancer classification. However, CNNs demand large training datasets that are generally not available in medical applications. 

Transfer learning (TL) is a common approach to counter the data deficiency problem. In TL, an architecture, pre-trained on a larger dataset for one application, is used in another related but data-deficient task. \cite{ref32} have highlighted the limitation of a small dataset in medical applications and the benefits of transfer learning in computer-aided detection.  A similar analysis has been presented by \cite{ref38} for nuclei classification in colorectal adenocarcinoma histopathology images. It is also emphasized that the depth of a pre-trained architecture and source data distribution impact performance in transfer learning. Similarly, \cite{ref40} and \cite{ref41} used transfer learning approach for the classification of HEp-2 cells, \cite{ref33} for lesion detection in mammograms, \cite{ref29} for cervical cell image classification, \cite{ref36} for skin cancer classification, and \cite{ref37} for the classification of histopathology images of brain tumor.

On a similar note, \cite{ref59} and \cite{ref60} employed a pretrained AlexNet for the prediction of different subtypes of ALL. \cite{Vogado2017} used a pretrained VGG-F architecture for feature extraction followed by PCA for features selection, and an ensemble of classifiers for the diagnosis of ALL. A similar methodology is adapted by \cite{Vogado2018}, wherein a pretrained CNN is used as a feature extractor followed by a feature selection mechanism, and an SVM for the classification of leukocytes. Again, a major downside of these methods is the utilization of small datasets for fine-tuning or evaluation purposes. For example, the datasets consist of only 310 images \citep{ref59}, 330 images \citep{ref60}, 108 images \citep{Vogado2017}, and 891 images \citep{Vogado2018}. Performance with such small datasets cannot be considered as the true performance indicator which can limit their deployment in practical scenarios. Also, the use of pre-trained architectures in medical applications require resizing of input images to the predefined input shape of these networks that may alter the morphology of input images impacting the performance of classifiers. Also, these architecture are designed to perform on a 1000-class ImageNet dataset of non-medical images. Such methods may not be suitable in medical applications for two reasons: 1) these are not trained on medical images, and 2) medical applications have, in general, relatively fewer classes.


In our previous work \citep{ref45}, we used a large dataset with 13407 cell images for ALL diagnosis and proposed a trainable stain deconvolutional layer (SD-Layer) at the front end of the CNN. The input RGB images were first transformed to the optical density (OD) space followed by stain color deconvolution to obtain stain quantity images. It was envisaged that the texture of stain quantity images captures the sub-microscopic cellular features, and hence, a CNN trained on these images would perform better compared to the one trained on RGB images. 
Indeed this architecture worked better with a CNN trained from scratch as compared to pre-trained architectures.

However, one of the most significant drawbacks was that the training data was not separated at the subject-level. The classifier was trained by pooling cell images of all the subjects into the respective
normal and cancer class. This pooling limited the performance of the classifier on prospective subjects' data because subject-specific characteristics 
also aid in class discrimination.  Moreover, in real life, sample cell images of new subjects are never available apriori for training. Thus, a realistic classifier would require training on some subjects' data and testing on completely unseen subjects' data. 
 
Hence, cell images were segmented again and the dataset was prepared in a curated manner by the expert oncologist at the subject-level with no intersection between the training subject-set and testing subject-set data. This strategy ensured that none of the cell images of a subject in the test data was used for training the classifier. This dataset is named C-NMC 2019 dataset \citep{ref49} and is publicly available at The Cancer Imaging Archive (TCIA). The dataset consists of a total of 15114 images collected from 118 subjects. As compared to the existing ALL datasets, C-NMC 2019 dataset contains a significantly large number of images that can prove helpful for developing a robust and deployable ALL diagnostic classifier. 

\begin{figure}[!ht]
\centering
\begin{subfigure}[b]{0.15\textwidth}
\includegraphics[width=1.5cm,height=1.5cm]{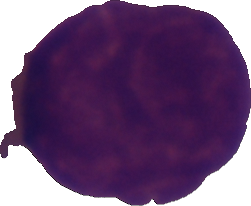}
\end{subfigure}
\begin{subfigure}[b]{0.15\textwidth}
\includegraphics[width=1.5cm,height=1.5cm]{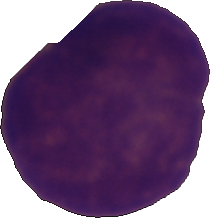}
\end{subfigure}
\begin{subfigure}[b]{0.15\textwidth}
\includegraphics[width=1.5cm,height=1.5cm]{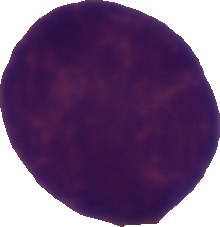}
\end{subfigure}
\begin{subfigure}[b]{0.15\textwidth}
\includegraphics[width=1.5cm,height=1.5cm]{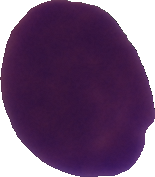}
\end{subfigure}

\begin{subfigure}[b]{0.15\textwidth}
\includegraphics[width=1.5cm,height=1.5cm]{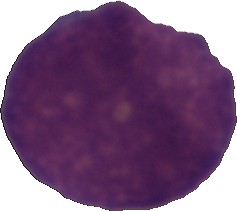}
\end{subfigure}
\begin{subfigure}[b]{0.15\textwidth}
\includegraphics[width=1.5cm,height=1.5cm]{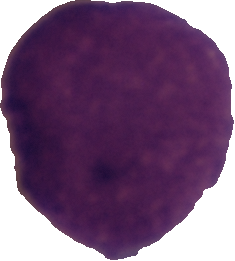}
\end{subfigure}
\begin{subfigure}[b]{0.15\textwidth}
\includegraphics[width=1.5cm,height=1.5cm]{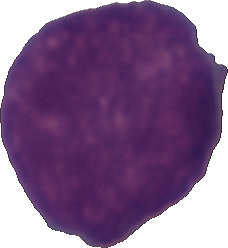}
\end{subfigure}
\begin{subfigure}[b]{0.15\textwidth}
\includegraphics[width=1.5cm,height=1.5cm]{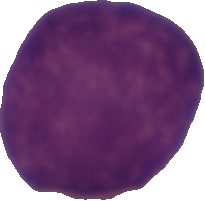}
\end{subfigure}
\captionsetup{justification=centering}
\caption{Sample images of different subjects from the cancer class (first row) and normal class (second row). This figure shows that the cell images are not distinguishable across classes or across subjects.}
\label{fig for samples images}
\end{figure}

In this paper, a novel deep learning architecture is proposed to differentiate malignant (cancer) immature  WBCs from normal (healthy) immature WBCs for computer-aided diagnosis of ALL. Instead of using transfer learning with a CNN architecture pre-trained on natural images, a CNN has been trained from scratch, as used by  \cite{ref27,ref34,ref30,ref35,ref45}. Because cell image features may differ widely from those of the natural images, a network trained from scratch on cell images may converge to a better solution . Although C-NMC 2019 dataset is quite large compared to the existing ALL datasets, still larger dataset would be required for training a very deep CNN architecture.  Hence, to curb this limitation, a compact CNN with fewer parameters is utilized. We have also exploited discrete cosine transform (DCT) to improve the classification performance. DCT finds use in numerous image processing applications including 
JPEG compression \citep{ref47}, brain tumor classification \citep{ref46}, and face recognition \citep{ref48}. We have used DCT to capture inter-class separability in the frequency domain because images of different classes in this dataset are not visually differentiable in the spatial domain, as can be verified from Fig.~\ref{fig for samples images}. 

Overall, the salient points of this work are as follows:
\begin{enumerate}	
    \item A novel deep learning architecture is proposed for the classification of cell images of ALL. The proposed architecture, namely, SDCT-AuxNet$^{\theta}$, is a 2-module framework and is end-to-end trainable. It consists of a compact CNN as the main classifier in one module and a Kernel SVM as the auxiliary classifier in the other one. 
    \item The input RGB images are first transformed to the OD space, and then stain color is deconvolved using a trainable stain deconvolutional layer (SD-Layer) appended at the front-end of the CNN to obtain stain quantity images. The texture of stain quantity images captures the sub-microscopic cellular features. Hence, a compact CNN trained on stain-deconvolved images in the OD space is envisaged to perform comparable or better than a deeper CNN trained on raw RGB images, overcoming the need of a big training dataset. 
    \item A DCT layer is used immediately after the SD-layer in the proposed CNN. The inspiration for training on the DCT features is derived from the fact that, conventionally, the spectroscopy-based method, namely flow cytometry, is used for ALL diagnosis.
       \item While the CNN classifier utilizes features via bilinear pooling employed before the fully connected layer, averaging across feature maps (spectral averaging) is used to create features for the auxiliary classifier. Bilinear pooling \citep{lin2015bilinear} helps in capturing fine-grained local feature interactions, whereas averaging across feature maps on the same set of input helps in capturing spectral counterpart that is used in parallel by the auxiliary classifier in the proposed architecture. This strategy allows different features to be used by the two classifiers.
    \item A novel test strategy is proposed that exploits both the classifiers for decision making using the confidence scores of their predicted class labels.
\end{enumerate}

Training and testing are carried out on a recently released in-house dataset of ALL cancer of 118 subjects \citep{ref49}. Benchmark in performance is achieved by comparing the results with the highest performance recorded on the leaderboard of the recently conducted challenge on this dataset at IEEE ISBI conference 2019 \citep{ref50}.

\section{Materials}
\vspace{-0.5em}
Microscopic images are captured from the bone marrow aspirate smears of subjects in the raw BMP format with a size of $2560 \times 1920$ pixels. Before imaging, smears are stained using Jenner-Giemsa stain for better visibility of B-type immature white blood cells (WBCs) under the microscope. The images are first normalized for stain color variability across images using the in-house pipeline \citep{ref53,ref57}. Next, cells of interest are marked by expert oncologists. These labeled cells are segmented using the in-house segmentation pipeline \citep{ref54,ref55}. Normal cell images are collected from subjects who did not suffer from cancer, while cancer cell images are obtained from subjects who are initially diagnosed with cancer. A dataset of 118 subjects, 49 healthy and 69 cancer subjects, is prepared at Laboratory Oncology, AIIMS, New Delhi, India. A waiver for written informed consent is obtained from the Ethics Committee of AIIMS, New Delhi, on this dataset for research purposes. One of the co-authors (RG) had access to the subject identifying information which is removed entirely from the image dataset before sharing of data with the other co-authors.

This dataset is publicly available at The Cancer Imaging Archive (TCIA) \citep{ref49} and displayed according to TCIA standard protocols \citep{ref56}. This dataset was also used in the medical imaging challenge, \textit{Classification of Normal vs. Malignant Cells in B-ALL White Blood Cancer Microscopic Images (C-NMC) 2019} \citep{ref50}, organized at IEEE International Symposium on Biomedical Imaging (ISBI), 2019 conference.

Data is prepared at the subject-level and is split into the training set and testing set. Training set consists of 8491 cancer cell images collected from 60 cancer subjects and 4037 normal cell images collected from 41 healthy subjects, with a total of 12528 images \citep{ref50}. This training data of 12528 cells is split into seven-folds for cross-validation, where the splits are created such that each fold contains an approximately equal proportion of images of both the classes. Also, the folds are prepared at the subject-level, i.e., all cells belonging to a subject are placed in the same fold. 

Test data contains a total of $2586$ cell images, collected from $9$ cancer subjects and $8$ healthy subjects \citep{ref50}. None of the cell images of these subjects are used for training the classifier and hence, results of the test data are unbiased. A detailed description of the training set and test set is provided in Table~\ref{data description}. Since original cell images are of different sizes, all images are zero-padded such that the centroid of every cell is at the center. After zero-padding, all cell images are made to the size of $350 \times 350$. 
 
\begin{table}[!ht]
\vspace{-.5em}
\begin{small}
\centering
\captionsetup{justification=centering}
\caption{Description of training and test data. \\ Training data is split in seven folds. All cells belonging to a subject are placed in the same fold and no two folds have cell images belonging to the same subject.}
\vspace{-1em}
\label{data description}
\begin{tabular}{|c|c|c|c|c|}
\hline
\multicolumn{5}{|c|}{{\textbf{Training Data}}} \\ \hline
{Fold} & \multicolumn{2}{c|}{{Cancer Class}} & \multicolumn{2}{c|}{{Normal Class}} \\ \hline
{} & {No. of subjects} & {No. of images} & {No. of subjects} & {No. of images} \\ \hline
1 & 13 & 1234 & 9 & 529 \\ \hline
2 & 9 & 1166 & 7 & 546 \\ \hline
3 & 9 & 1269 & 6 & 516 \\ \hline
4 & 6 & 1275 & 4 & 624 \\ \hline
5 & 9 & 1135 & 6 & 618 \\ \hline
6 & 6 & 1197 & 4 & 603 \\ \hline
7 & 8 & 1215 & 5 & 601 \\ \hline
{Total} & {60} & {8491} & {41} & {4037} \\ \hline
\multicolumn{2}{|c|}{{Total no. of subjects}} & 101 & {Total no. of images} & 12528 \\ \hline
\multicolumn{5}{|c|}{{\textbf{Test Data}}} \\ \hline
\multicolumn{3}{|c|}{{Cancer Subjects}} & \multicolumn{2}{c|}{{Healthy Subjects}} \\ \hline
\multicolumn{3}{|c|}{9}  & \multicolumn{2}{|c|}{8}  \\ \hline
\multicolumn{3}{|c|}{{Total no. of subjects}} & \multicolumn{2}{|c|} {{17}} \\ \hline
\multicolumn{3}{|c|}{{Total no. of cell images in the test data}} & \multicolumn{2}{|c|} {2586} \\ \hline
\end{tabular}
\end{small}
\end{table}

\section{Methods}
In this section, the proposed SDCT-AuxNet$^\theta$ architecture and other variants are presented in detail.  
\subsection{Base Network} 
\begin{figure*}[!t]
\centering
\centerline{
\includegraphics[scale=0.3]
{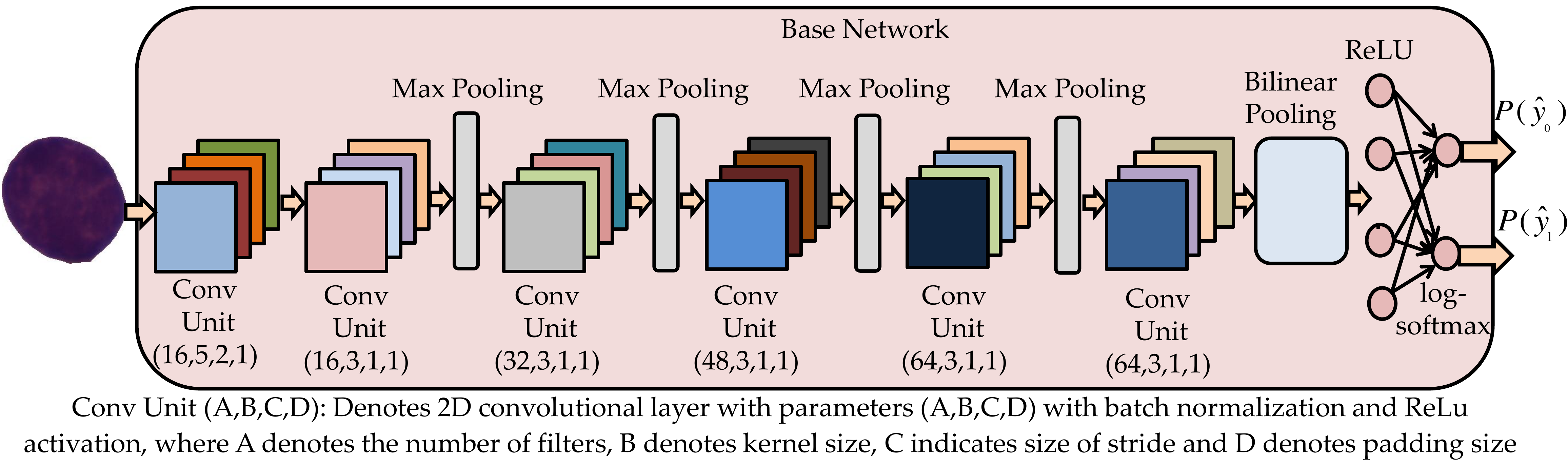}}
\captionsetup{justification=centering}
\caption{Base network. This network is used as the building block in SD-Net as well as in the proposed architectures.}

\label{base_network}
\end{figure*}

The size of the input dataset is a crucial factor in designing a CNN because an architecture with large learnable parameters is prone to overfitting, while a CNN with too few learnable parameters tends to underfit. Hence, it is essential to design an architecture with an adequate number of tunable parameters. Moreover, transfer learning is not appropriate because all well-known architectures are trained on natural images, while this dataset specifically contains blood cell images. Recently, a compact CNN with fewer parameters was designed on this ALL dataset  \citep{SimmiLeukoNet} as shown in Fig. \ref{base_network}. This architecture has a convolutional layer with stride two at the front-end followed by five convolutional layers with batch-norm, maxpool, and ReLu. \cite{lin2015bilinear} introduced the idea of bilinear pooling, wherein the outer product of the feature vectors obtained through two different CNNs is connected to a common output layer. This idea is adopted in this base network, and the outer product of the feature vector with itself is connected to the classification layer. 

The output of bilinear pooling is passed through a linear layer followed by the log-softmax layer. This network is referred to as the \textit{base network}. Here, \textit{base network} is used as the building block in SD-Net architecture as well as in the following proposed architectures. A comparison of the number of learnable parameters of the  \textit{base network} with some well-known architectures used in the transfer learning approach is shown in Table \ref{parameters comparison}. It is evident that the \textit{base Network} has fewer parameters compared to those in the existing architectures. Hence, base network is suitable for this dataset that is, although very large compared to existing public ALL datasets, is still relatively small compared to the ImageNet dataset on which often-used architectures are trained. 

\begin{table*}[!t]
\begin{small}
\captionsetup{justification=centering}
\caption{Comparison on the number of learnable parameters. \\
The number of parameters in the \textit{base network} is significantly less as compared to existing architectures used commonly in transfer learning approach.}
\vspace{-1em}
\label{parameters comparison}
\centerline{
\begin{tabular}{|c|c|c|c|c|c|c|}
\hline
{Architecture} & AlexNet & VGG16 & ResNet-18 & DenseNet-161 & Inception-v3 & Base Network \\ \hline
{Parameters} & 61100840 & 138357544 & 11689512 & 28681000 & 27161264 & \textbf{95362} \\ \hline
\end{tabular}%
}
Note: Least number of parameters are specified in bold.
\end{small}
\end{table*}

\subsection{SD-Net Architecture}
\begin{figure*}[!ht]
\centerline{
\includegraphics[scale=0.31]
{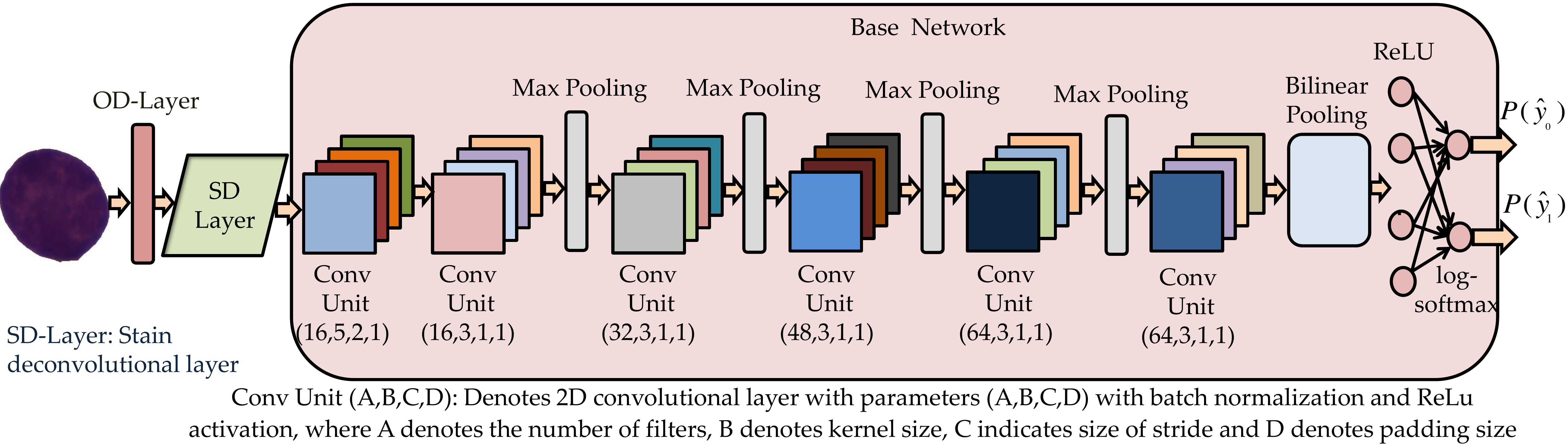}}
\captionsetup{justification=centering}
\caption{SD-Net architecture. SD-layer deconvolves stain color vector in the OD space and provides stain quantity images, i.e., images that contain the quantification of the absorbed stain, a characteristic of the tissue.}

\label{SD-Net_model}
\end{figure*}

In this architecture, SD-layer \citep{ref45} is stacked at the front-end of the \textit{base network}. It is, hereby, named as SD-Net (refer to Fig.~\ref{SD-Net_model}).  Images obtained from the SD-layer are the stain deconvolved images in the optical density (OD) space that are fed into the \textit{base network}.  The deconvolution of stain in the OD colorspace allows the computation of pixel stain quantities that enable the network to see tissue-specific stain absorption quantities. It has been shown by \cite{ref45} that the introduction of the SD-layer as the first layer leads to better performance of the classifier. Motivated by this, SD-Net utilizes SD-layer at the front-end and hence, works in the OD-space rather than the RGB space. 
SD-Net is also used as the baseline classifier in experiments.

\begin{figure}[!ht]
\centerline{
\includegraphics[scale=.7]
{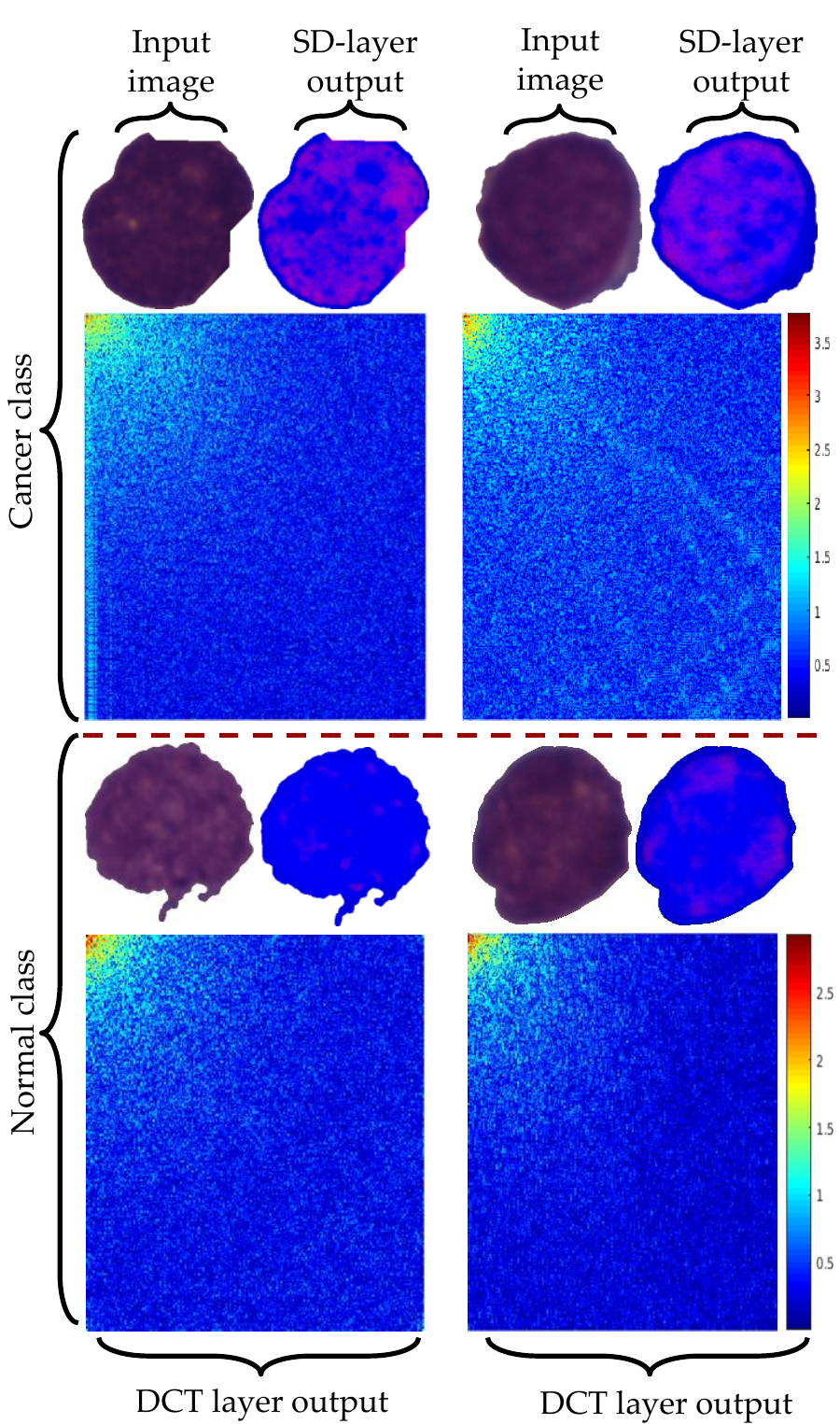}}
\captionsetup{justification=centering}
\caption{Impact of DCT-layer on the output images of SD-layer. Normal cell images are sparsified more  compared to cancer cell images with the DCT-layer.}

\label{dct output}
\end{figure}

\subsection{Proposed SDCT-Net Architecture}
We build upon the SD-Net to achieve better class discrimination capability. As seen in Fig. \ref{fig for samples images}, images of the two classes are visually indistinguishable. In such a scenario, it is more intuitive to work in another domain that is more helpful in capturing class-specific characteristics. The assumption is that each class would have some distinct properties that can be highlighted by applying a specific transform. For images, one such property is sparsity. A signal $\textbf{x}$ is $r-$ sparse if only $r$ elements are non-zero. In general,  compressibility, which is less strict condition than sparsity, is used frequently. The magnitude of sorted coefficients of a compressible signal decay quickly. Images are generally compressible in the discrete cosine transform (DCT) domain. The famous JPEG compression uses a compression technique based on DCT. This property of compressibility of DCT is used in many other applications, including denoising and compressive sensing based reconstruction \citep{ref52}.  
 
 In general, images have different sparsification levels, depending upon their internal structure. This understanding can be extended to images of different classes as well, i.e., images of different classes will be sparsified differently. To analyze this effect,  we appended a DCT layer after the SD-layer. The output of the DCT-layer is given as:  
\begin{equation}
\textbf{D}^j=\text{log}_{10}\big(\textbf{I}+ |\psi(\textbf{Z}^j)|\big),\ \ \ \ \ \ j=0,1,2,
\end{equation}
where $\psi(.)$ denotes the 2D-DCT operator, $\textbf{Z}^j$ denotes the  $j^\text{th}$ feature map input to the DCT layer, and \textbf{I} denotes a matrix with `1' at each position of the matrix. This `1' is added to the DCT output at every pixel before logarithm so that the resulting image \textbf{D} contains only positive quantities. Since the dynamic range of the DCT output image is broad, ``log" acts as the normalization function and allows the contribution from all pixels to be exploited by the following layers.

\begin{figure*}[!ht]
\centering
\centerline{
\includegraphics[scale=0.31]
{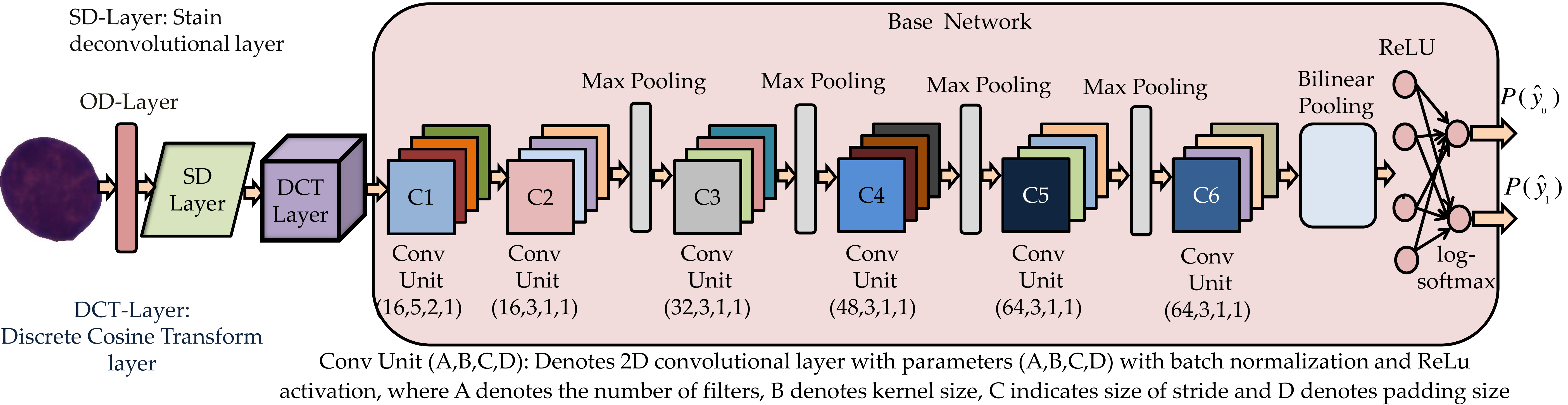}}
\captionsetup{justification=centering}
\caption{Proposed SDCT-Net. DCT layer is appended after SD-layer which helps to learn class discriminative features by inducing different sparsity in cancer and normal cell images.}

\label{cnn_proposed_model}
\end{figure*}

The feature maps/images obtained at the output of the SD-layer are passed to the above explained DCT-layer. The impact of DCT-layer on these feature maps is shown in Fig.~\ref{dct output}. Interestingly, it is observed that the normal cell images are sparsified more by DCT-layer compared to the cancer cell images. The dynamic range of DCT coefficients of the two classes is also different. This observation can be exploited by the following \textit{base network} to enhance its classification performance. This architecture with DCT layer between the SD-layer and the \textit{base network} is shown in Fig.~\ref{cnn_proposed_model} and is, hereby, named as SDCT-Net.  The size of the input image and feature maps at different layers of SDCT-Net are shown in Table~\ref{output size}. This architecture is a sub-part of the proposed architecture $\text{SDCT-AuxNet}^\theta$, described next.

\begin{table}[!ht]
\begin{small}
\centering
\caption{Size of input image, filters, and output feature maps at different layers of SDCT-Net}
\vspace{-1em}
\begin{tabular}{c c c c c}
\hline
Layer           & Layer Name & Filter Size & Input Size    & Size of Feature Map Output   \\ \hline
SD-layer & -      & - &$3 \times 350 \times 350$  & $3 \times 350 \times 350$   \\
DCT-layer & -      & - & $3 \times 350 \times 350$  & $3 \times 350 \times 350$   \\
First Conv layer  & C1& $5 \times 5$       & $3 \times 350 \times 350$  & $16 \times 174 \times 174$   \\
Second Conv layer         & C2 & $3 \times 3$       & $16 \times 174 \times 174$ & $16 \times 174 \times 174$    \\
First Pooling         & - & $2 \times 2$       & $16 \times 174 \times 174$ & $16 \times 87 \times 87$    \\
Third Conv layer & C3 & $3 \times 3$       & $16 \times 87 \times 87$ & $32 \times 87 \times 87$     \\
Second Pooling         & -& $2 \times 2$       & $32 \times 87 \times 87$ & $32 \times 43 \times 43$    \\
Fourth Conv layer & C4 & $3 \times 3$       & $32 \times 43 \times 43$   & $48 \times 43 \times 43$  \\
Third Pooling         & -& $2 \times 2$       & $48 \times 43 \times 43$ & $48 \times 21 \times 21$    \\
Fifth Conv layer  & C5& $3 \times 3$       &$ 48 \times 21 \times 21$   & $64 \times 21 \times 21$    \\
Fourth Pooling         &-&  $2 \times 2$       & $64\times 21 \times 21$ & $64 \times 10 \times 10$    \\
Sixth Conv layer &C6&  $3 \times 3$       & $64 \times 10 \times 10$   & $64 \times 10 \times 10$     \\
Bilinear Pooling & -       & $64 \times 10 \times 10$   & $4096\times 1$     \\
Linear          &    -         & $4096\times 1$    & $2\times 1$               \\ \hline
\end{tabular}
\label{output size}
\end{small}
\end{table}

\subsection{Proposed SDCT-AuxNet$^{\theta}$}
\begin{figure}[!ht]
\centering
\centerline{
\includegraphics[scale=.61]
{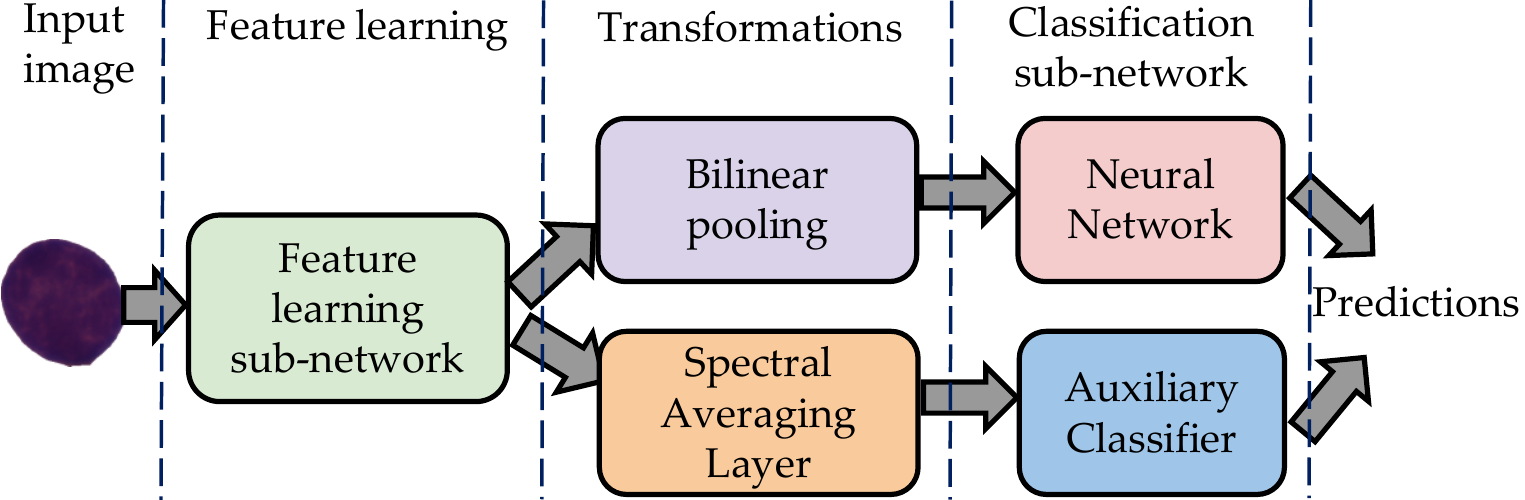}}
\captionsetup{justification=centering}
\caption{An overview of the proposed SDCT-AuxNet$^{\theta}$. The SDCT-AuxNet$^{\theta}$ can be interpreted as feature learning sub-network with two different classifiers, one classifier operating on the features obtained after bilinear pooling, and other classifier using the features obtained through spectral averaging.}
\label{network subpart}
\end{figure}

Next, we propose a novel framework that consists of SDCT-Net and an additional auxiliary classifier. An overview of this architecture is shown via a block diagram in Fig.~\ref{network subpart}. SDCT-AuxNet$^{\theta}$ has two sub-networks: i) features learning sub-network and ii) classification sub-network. In general, learned features are applied to a neural network after applying some transformations. But, SDCT-AuxNet$^{\theta}$ processes 
the feature maps obtained from the feature learning sub-network through two different transformations: 1) bilinear pooling that feeds to the neural network for decision making and 2) spectral averaging that carries out averaging across all feature maps at every pixel position and feeds the output to the auxiliary classifier. 

\subsubsection{Training Methodology for the proposed SDCT-AuxNet$^{\theta}$} 
\begin{figure*}[!ht]
\centering
\centerline{
\includegraphics[scale=.28, trim=0 20 0 35]
{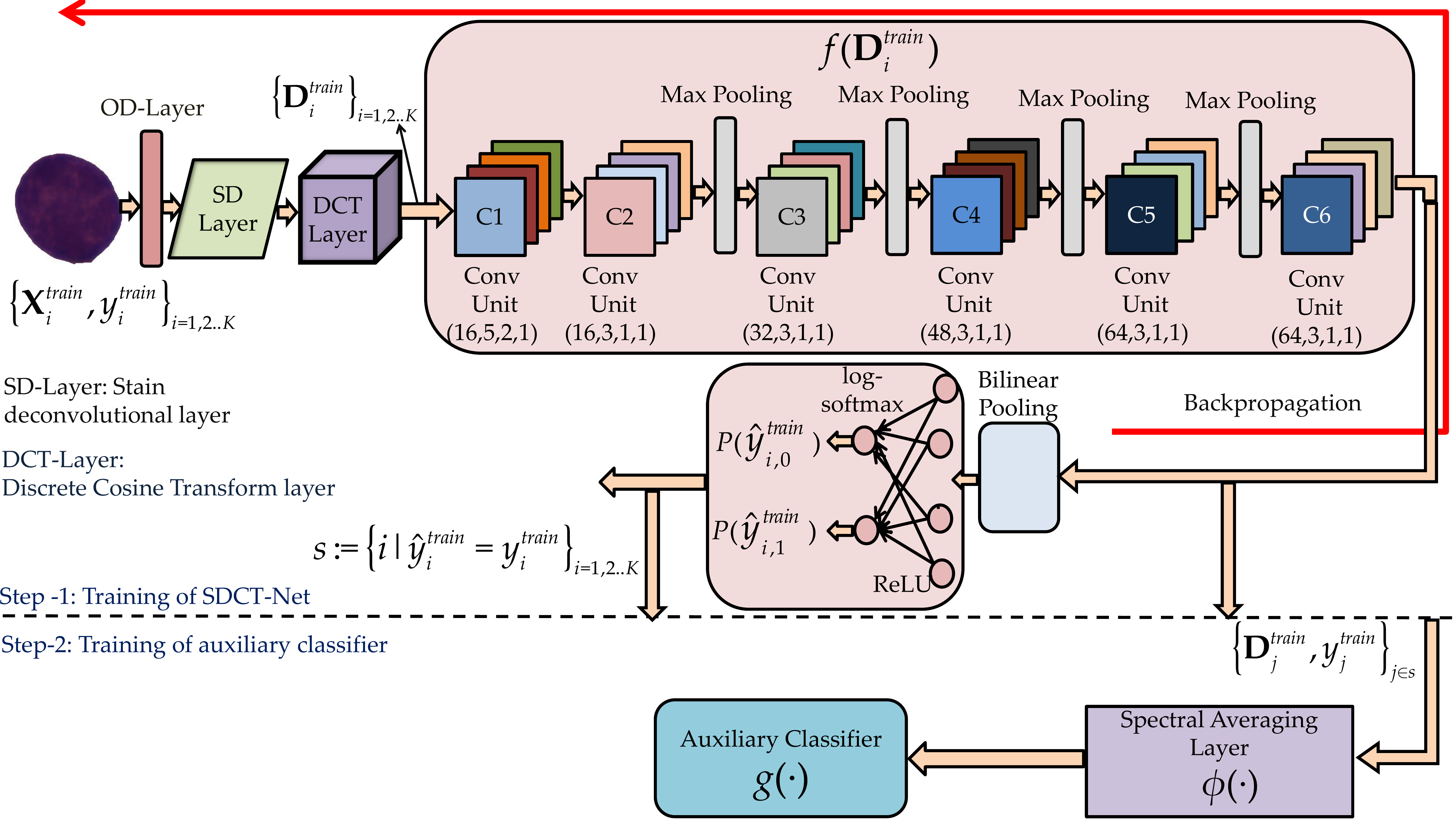}}
\captionsetup{justification=centering}
\caption{Two-steps training of SDCT-AuxNet$^{\theta}$. In Step-1, SDCT-Net is trained in an end-to-end fashion on the entire training set containing $K$ training images with indices $i=1,2,...,K$. In Step-2, correctly classified training images by SDCT-Net are used to train an auxiliary classifier. 
}
\label{training_strategy}
\end{figure*}

The training procedure of SDCT-AuxNet$^{\theta}$ is depicted with more clarity in Fig.~\ref{training_strategy}. Although two different classifiers are used (main CNN classifier and auxiliary classifier), they are not trained simultaneously. Let $\{\textbf{X}_i^{train}\}_{i=1,2,...,K}$ with size $m \times n \times 3 $ represents a set of $K$ input images and $\{y_i^{train}\}_{i=1,2,...,K}$ represents true labels of the respective images. In Step-1, SDCT-Net (feature learning sub-network + neural network) is trained in an end-to-end fashion by minimizing the negative log likelihood loss function. Let $P(\hat{y}_{i,0}^{train})$ and $P(\hat{y}_{i,1}^{train})$ denote the probabilities that the $i^{th}$ training image belongs to normal class (label $0$) and 

\noindent cancer class (label $1$), respectively. The predicted label is given as
\begin{align}x
  \hat{y}_i^{train}&=1\ \ \ \ \ \ \ \ \ \text{if}\ \ \ \ {P}(\hat{y}_{i,1}^{train})> (\hat{y}_{i,0}^{train}),\nonumber \\
  &= 0 \, \, \, \,  \,\, \text{\,\,\,\,\,Otherwise}
\label{eq.4}  
\end{align}
Once SDCT-Net is trained, we predict labels of all the training images using this trained SDCT-Net. We define $s$ as a set of indices of training images for which SDCT-Net made the correct predictions, i.e.,
\begin{equation}
s:=\{i| \hat{y}_i^{train}={y}_i^{train}\}, \,\,\, \text{for   } \,\,\, i=1,2,...,K.
\end{equation}

In Step-2, images $\{\textbf{X}_j^{train}\}_{j \in s}$ are utilized to train the  auxiliary classifier. These images are first processed by the feature learning sub-network followed by the spectral averaging layer ($\phi(\cdot)$).
This is to note that only the images, whose labels are correctly predicted by the main CNN classifier (SDCT-Net), are used to train the auxiliary classifier. Also, the auxiliary classifier is not used to update the parameters of the feature learning sub-network. The strategy is to exploit two different classifiers and yet train the architecture in an end-to-end fashion. The intuition behind this approach is simple. Different transformations applied to learned features will capture different information and hence, two different but coupled classifiers trained on two distinct sets of features of the same images may perform better.

\subsubsection{Testing Methodology for proposed SDCT-AuxNet$^{\theta}$}
\label{section for testing strategy }
Once the SDCT-AuxNet$^{\theta}$ is trained, we use the following strategy to predict the label of an input test image 
$\textbf{X}^{test}$. 
\begin{figure*}[!ht]
\centering
\centerline{
\includegraphics[scale=0.28, trim=0 10 0 0]
{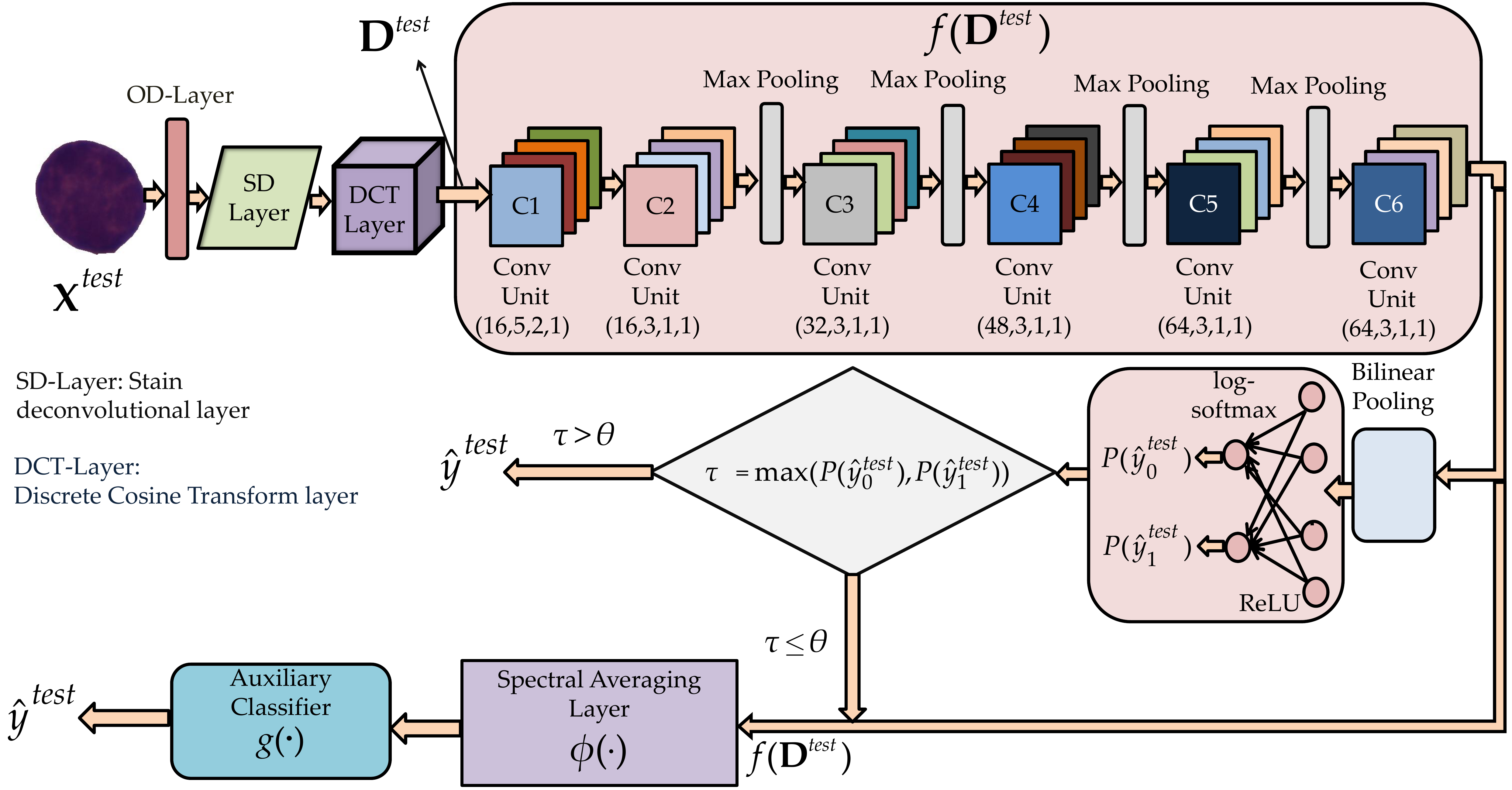}}
\captionsetup{justification=centering}
\caption{Testing strategy of the proposed SDCT-AuxNet$^{\theta}$. For each sample, probability of SDCT-Net decision label is compared against the confidence probability. If the probability of the predicted label by SDCT-Net is less than the confidence probability, it is predicted again using the auxiliary classifier.}
\label{testing_strategy}
\end{figure*}
First, a prediction is made using the main CNN classifier of the SDCT-AuxNet$^{\theta}$ via \eqref{eq.4}. Let $P(\hat{y}_{0}^{test})$ and $P(\hat{y}_{1}^{test})$ denote the prediction probabilities that the test image belongs to normal class (label $0$) and cancer class (label $1$), respectively. These prediction probabilities on the test image represent the confidence of the main CNN classifier in making the predictions for the respective labels. We introduce a confidence probability $\theta \in [0,1]$ as a user-defined variable. We assume the prediction from the CNN classifier to be correct if its predicted label's probability is higher than $\theta$. For more clarity, let us define 
\begin{equation}
\tau = max \left(P(\hat{y}_0^{test}),P(\hat{y}_1^{test})\right).
\end{equation}
This parameter $\tau$ is compared with the user-defined confidence probability $\theta$ and the decision is made as below:
\begin{equation}
\hat{y}^{test}=c\ \ \ \text{if}\ \ \tau> \theta \text{\,\, and \,\,} P\left(\hat{y}_{c}^{test}\right)> P\left(\hat{y}_{1-c}^{test}\right), \ \ \text{\, where}\ \ c=0,1.
\end{equation}

However, if $\tau$ less than $\theta$, we use the auxiliary classifier to predict the label of that image and consider that to be the final prediction. 

To predict a label using the auxiliary classifier, features of the input image obtained from \textit{feature learning sub-network} are given to the spectral averaging layer, and the output of this layer is fed to the auxiliary classifier. The prediction of the auxiliary classifier is given as:
\begin{equation}
\hat{y}^{test}=g(\phi(f(\textbf{D}^{test}))),
\end{equation}
where $\phi(f(\textbf{D}^{test}))$ is the output of spectral averaging layer and $f(\textbf{D}^{test})$ are the feature maps obtained from $f(\cdot)$. This process is demonstrated in Fig.~\ref{testing_strategy}. Using this approach, we take into consideration the impact of two different classifiers, a compact CNN classifier and an auxiliary classifier. Since we are making predictions with two different classifiers using different set of features, we hypothesize that the incorrect prediction of one classifier can be corrected by the other classifier.  

\section{Results}
In this section, we analyze all the above discussed architectures and also compare their performances. 
\subsection{System Configuration}
A forty core system with 126 GB RAM is used to train and test all the networks. We have also used an Nvidia GTX 1080 Ti GPU to boost up the training and testing process. Also, PyTorch deep learning library is used to implement all the architectures. We have used stochastic gradient descent (SGD) optimizer with an initial learning rate of 0.001 and a batch size of 64. A total of 130 epochs are used for training the networks, during which the learning rate is dropped regularly by a factor of 10.

\subsection{Performance Metrics}
We have used $F_1$ score, weighted $F_1$ score and balanced accuracy to evaluate the performance of classifiers. Weighted $F_1$ score and balanced accuracy are reported to take into account the class imbalance problem of the data. Further, we report $F_1$ score on each class to show class-wise performance. $F_1$ score is defined mathematically as
\begin{equation}
F_1=2\times \frac{precision\times recall}{precision + recall}.
\end{equation} 
For binary data, weighted $F_1$ score is defined as
 \begin{equation}
 WF_1=\frac{\sum_{j=0}^1{n(c_j)F_1(c_j)}}{N},
 \end{equation}
where $F_1(c_j)$ is the $F_1$ score of $j^{th}$ class, $n(c_j)$ is the number of samples in $j^{th}$ class, and $N$ is total number of samples. Finally, balanced accuracy is defined as
\begin{equation}
BAC=\frac{recall+specificity}{2}.
\end{equation}
Precision is the ratio of correctly detected positive samples to  all predicted positive samples. Recall is defined as the ratio of correctly predicted positive samples to all positive samples in the ground truth. Specificity denotes the ratio of correctly classified negative samples to all negative samples in the ground truth.
Recall reflects the classifier's ability to correctly diagnose the disease, while specificity shows the classifier's effectiveness in discarding the healthy subjects. For an ideal classifier, the value of precision, recall, and specificity is 1.
\subsection{Classifiers' Training and Testing Information}
\label{majorityvoting}
For training and testing the networks, we used seven-fold cross-validation strategy. The training set is divided into seven-folds. At any time, six folds are used for training and the remaining one fold is used for validation. Therefore, for every architecture (SD-Net, SDCT-Net, or SDCT-AuxNet$^{\theta}$), we have a total of seven trained models: one for each different validation fold. To make the final prediction on the new test sample $\textbf{X}_{i}^{test}$, we note the individual prediction from each model as
\begin{equation}
\hat{y}_{i}^{test,m}=\gamma_m(\textbf{X}_{i}^{test})\ \ \ \ \ \ \ \ \ m=1,2,...,7, \text{\, and \,\,} i=1,2,...,N,
\end{equation}
where $\gamma_m(.)$ represents the output of the model trained using the $m^{th}$ fold as the validation set and $N$ denotes the total number of test samples. 
The final prediction for each $i^{\text{th}}$ test sample is given by majority vote as:
\begin{equation}
\hat{y}_{i}^{test}=mode \{\hat{y}_{i}^{test,1},\hat{y}_{i}^{test,2},...,\hat{y}_{i}^{test,7}\}.
\end{equation}
To summarize, the label predicted by the majority of the models for an architecture (SD-Net, SDCT-Net, or SDCT-AuxNet$^{\theta}$) is chosen as its final prediction. 

Another issue that needs to be addressed during the training is class imbalance because training using an imbalanced dataset may result in a biased classifier. From Table~\ref{data description}, it is noted that the training set is skewed towards the cancer class as the number of cancer cell images is approximately twice the total number of normal cell images. To resolve this, we have used the \textit{oversampling} strategy in which images of the minority class are duplicated, such that the total sample count of both the classes is the same.

Further, to enhance the performance, we have used augmentation techniques that simulate the conditions of real-life data preparation. In general, three types of augmentations are natural for these cell images. First, it is observed that 3-dimensional cells get sheared to some extent on the 2-dimensional slide when cells are pushed between two glass slides during the microscopic slide preparation. Secondly, cells would fall on the slide at varying orientations and hence, would be captured with different rotated angles. Thirdly, there can be variations in the sharpness of the images because the manual focusing of the microscope lens was carried out while capturing the images. Thus, to mimic these effects during the training process, shear augmentation is used within a range of [-20, 20]; augmentation of random rotation, horizontal flip, and vertical flip is used; and Gaussian blur augmentation is used within a range of [0.0, 0.75]. The same augmentations are used with all architectures: SD-Net, SDCT-Net, and SDCT-AuxNet$^{\theta}$. 

\begin{figure*}[!ht]
\centerline{
\begin{subfigure}[b]{0.4\textwidth}
\includegraphics[scale=0.318]{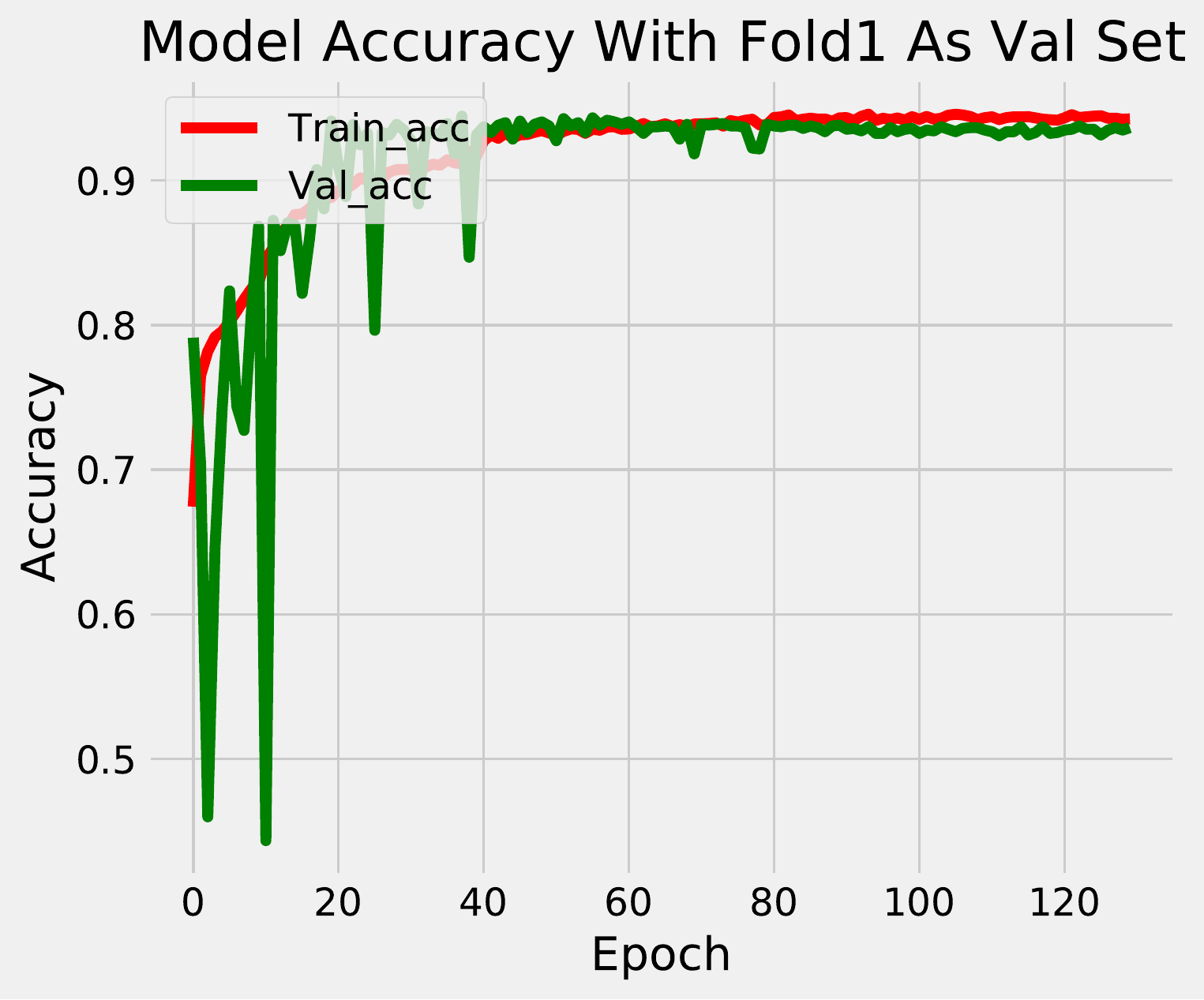}
\captionsetup{justification=centering}
\end{subfigure}
\begin{subfigure}[b]{0.4\textwidth}
\includegraphics[scale=0.318]{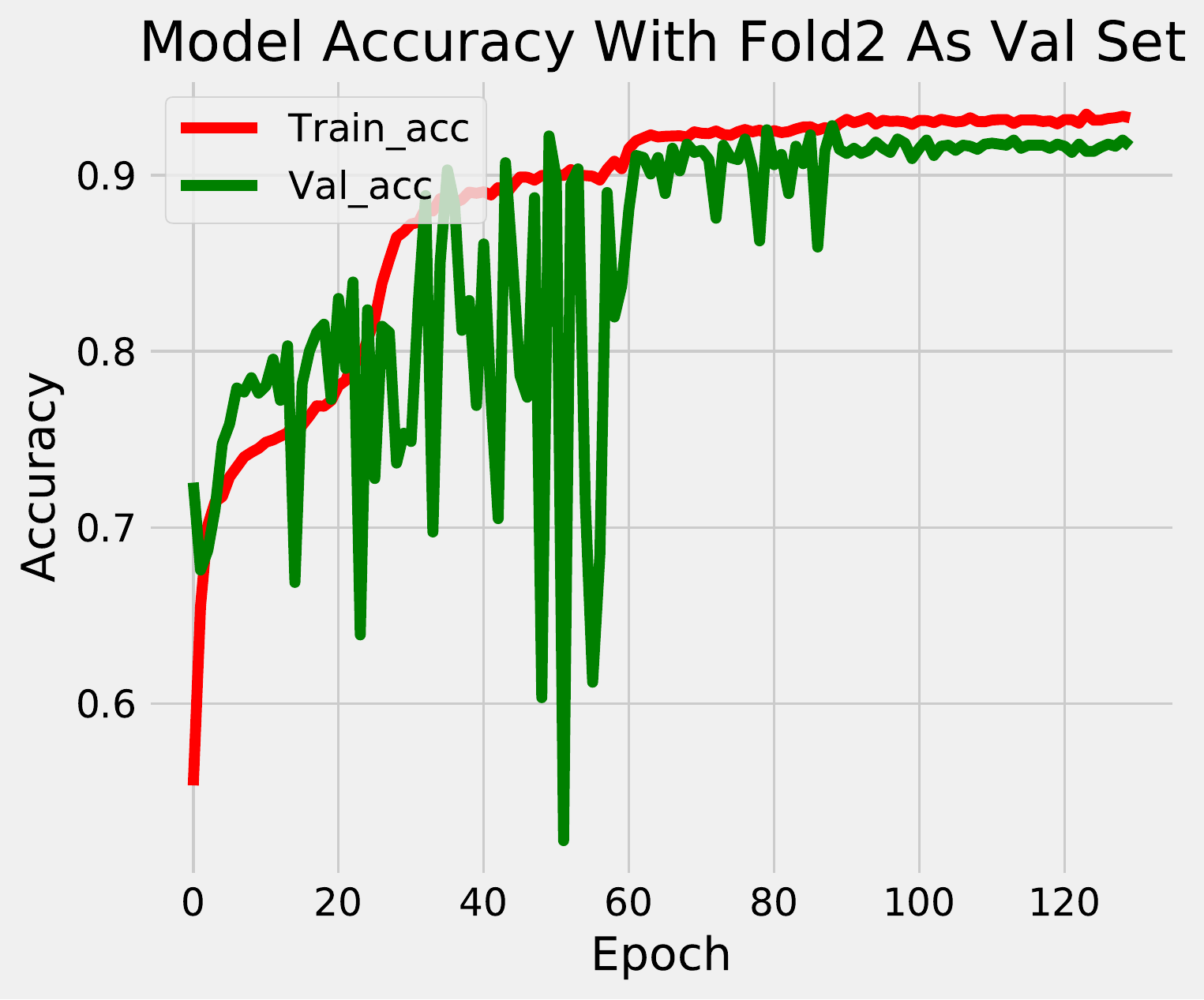}
\captionsetup{justification=centering}
\end{subfigure}
\begin{subfigure}[b]{0.4\textwidth}
\includegraphics[scale=0.318]{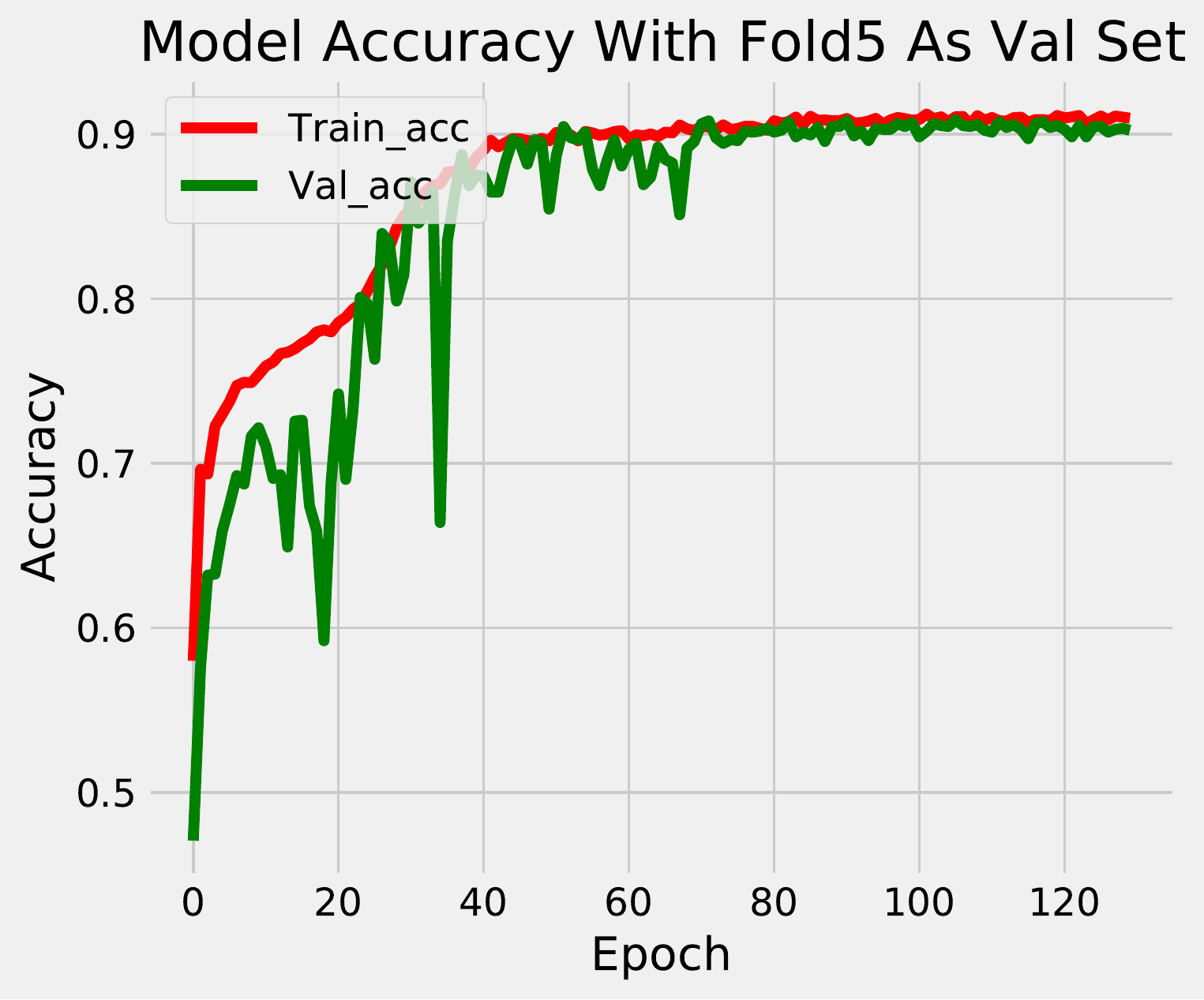}
\captionsetup{justification=centering}
\end{subfigure}}
\centerline{
\begin{subfigure}[b]{0.4\textwidth}
\includegraphics[scale=0.32]{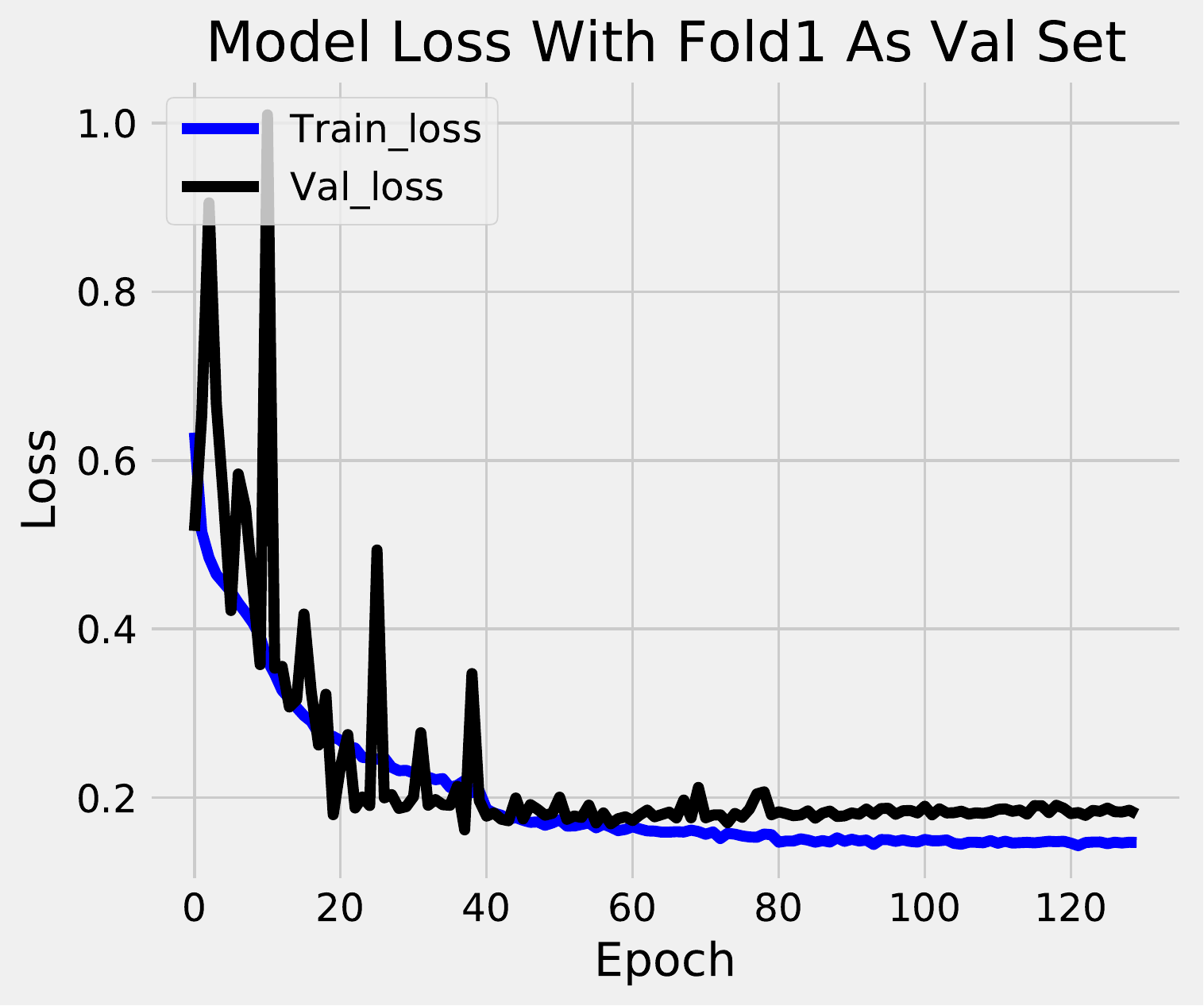}
\captionsetup{justification=centering}
\end{subfigure}
\begin{subfigure}[b]{0.4\textwidth}
\includegraphics[scale=0.32]{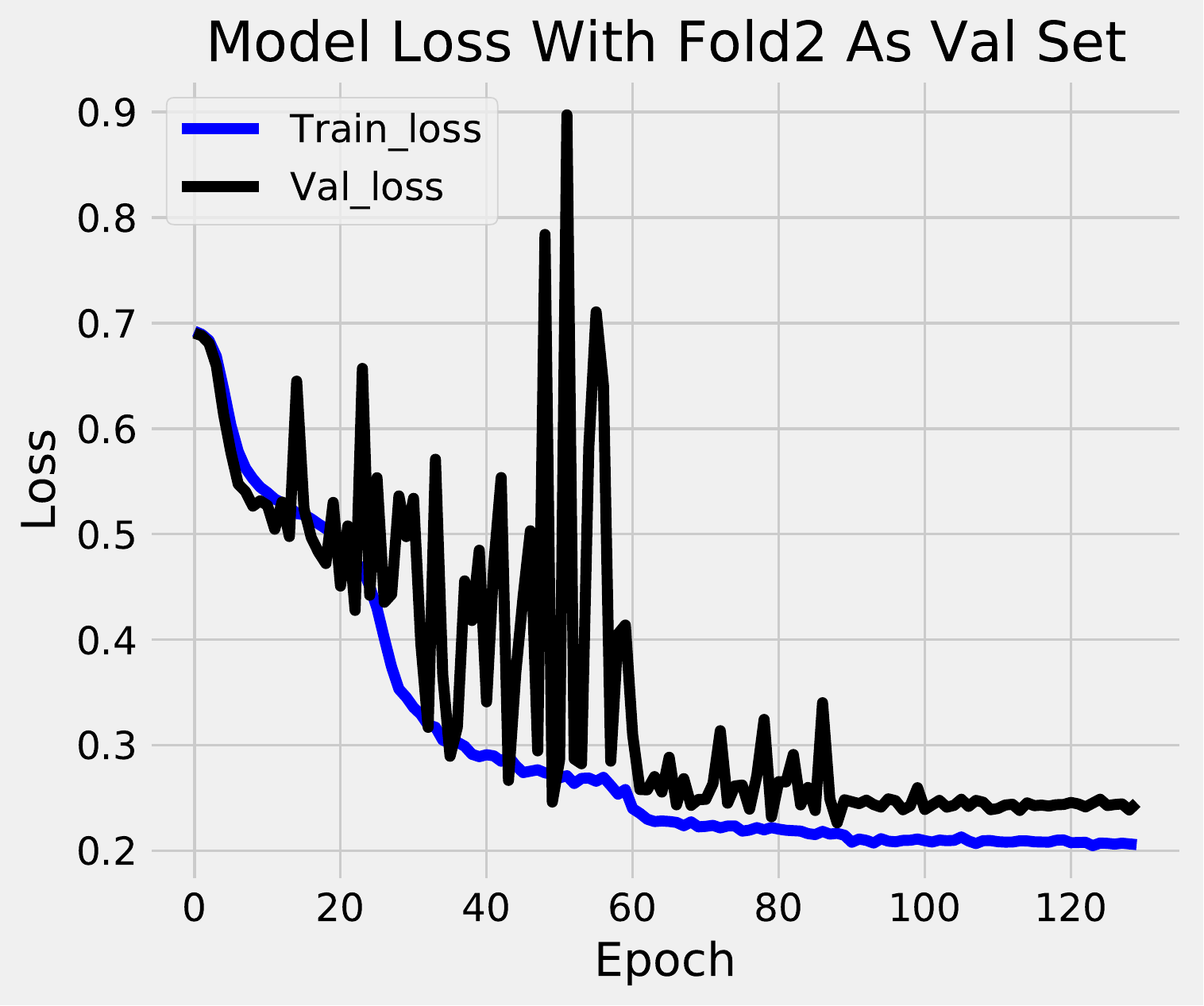}
\captionsetup{justification=centering}
\end{subfigure}
\begin{subfigure}[b]{0.4\textwidth}
\includegraphics[scale=0.32]{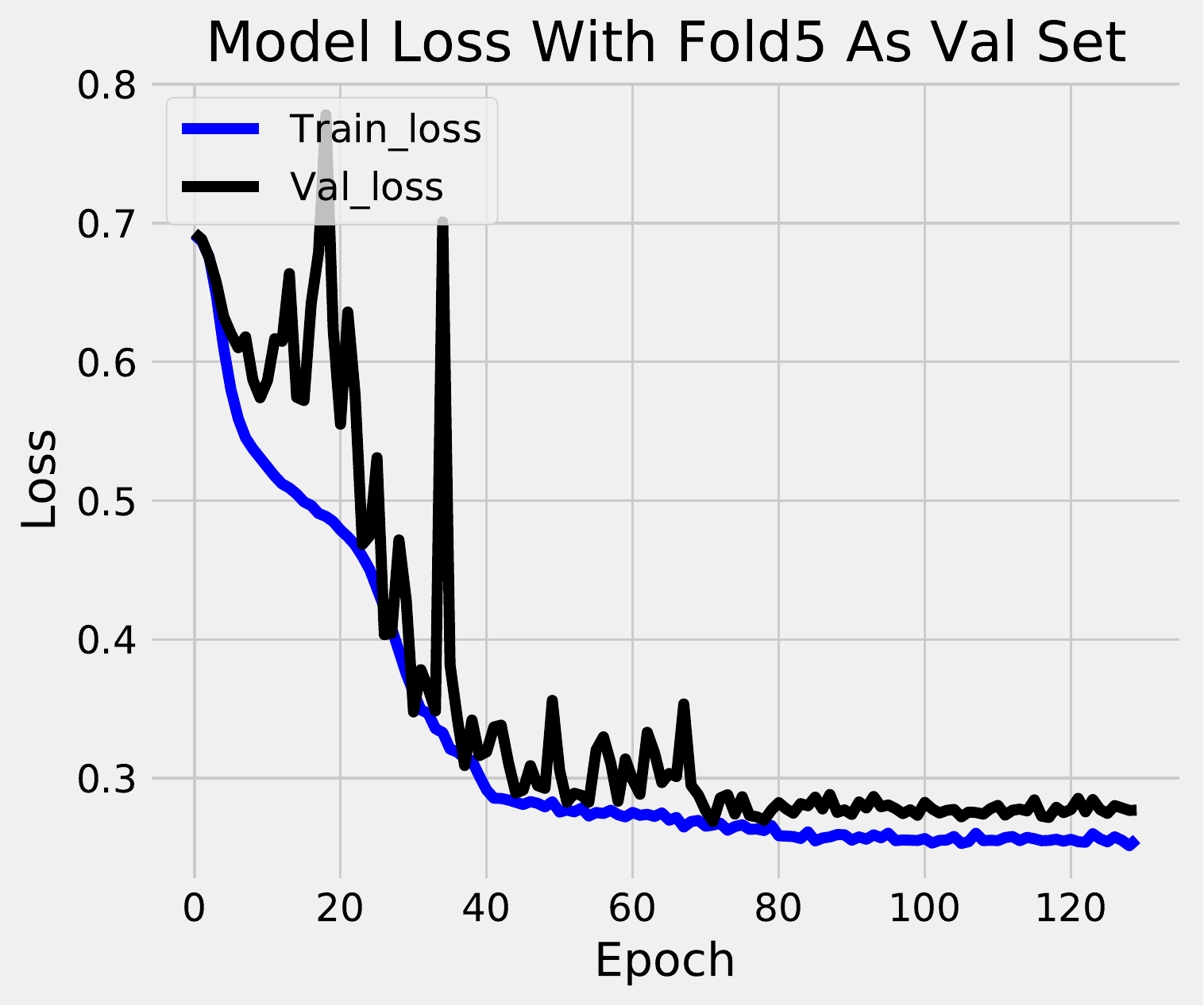}
\captionsetup{justification=centering}
\end{subfigure}}
\captionsetup{justification=centering}
\caption{Training curves of SDCT-Net with three different folds as the validation sets.}
\label{training curves}
\end{figure*}

Training curves of SDCT-Net are shown in Fig. \ref{training curves}. Graphs show the accuracy and loss of SDCT-Net trained with different folds as the validation sets. It is observed from the graphs that SDCT-Net converged in 130 epochs. Also, SDCT-Net is able to generalize well because the performance on training and validation sets is similar. Similar observation is noted for SDCT-AuxNet$^{\theta}$.


\subsection{Comparison of SD-Net and SDCT-Net}
Performance comparison of SD-Net and SDCT-Net in terms of $F_1$ score, $WF_1$ score, and $BAC$ is shown in Table \ref{comparison of SD-Net and SDCT-Net}. Results are tabulated using the individual models (Model1-Model7), i.e., models trained with a particular fold as the validation set and also using the majority voting as discussed in Section~\ref{majorityvoting}. From Table~\ref{comparison of SD-Net and SDCT-Net}, it is observed that SDCT-Net performs better than SD-Net on all three metrics for most of the models (Model1-Model7). Finally, with a majority voting on the decision of Model1 to Model7 SDCT-Net outperforms SD-Net with a good margin. The margin is maximum for $F_1$ score of normal class with SDCT-Net at  $0.862$ and SD-Net at $0.820$. SDCT-Net achieved $WF_1$ score of $0.908$ which is comparatively high compared to $WF_1$ score of $0.873$ with SD-Net. BAC with SDCT-Net and  SD-Net is $0.911$ and  $0.885$, respectively, showing superior performance of SDCT-Net. 
\begin{table}[!ht]
\vspace{1em}
\begin{small}
\centering
\captionsetup{justification=centering}
\caption{Comparison of SD-Net  and SDCT-Net. SDCT-Net outperforms SD-Net on all performance metrics. Model1 represents the model trained with fold1 as the validation set and remaining folds as the training set. Similar notation applies to Model2 to Model7. Results are tabulated on the test set of 2586 images. Best results are depicted in boldcase.}
\vspace{-1em}
\label{comparison of SD-Net and SDCT-Net}
\centerline{
\begin{tabular}{c c c c c c c c c}
\hline
\multicolumn{9}{c}{{$F_1$ Score for Normal Class}} \\ \hline 
{Architecture\textbackslash Model} & {Model1} & {Model2} & {Model3} & {Model4} & {Model5} & {Model6} & {Model7} & {Majority Voting} \\ \hline
{SD-Net} & 0.790 & 0.828 & 0.822 & 0.771 & 0.788 & {0.803} & {0.818} & 0.820 \\ 
{SDCT-Net} & {0.813} & {0.855} & {0.836} & {0.856} & {0.822} & 0.774 & {0.825} & \textbf{0.862} \\ \hline 
\multicolumn{9}{c}{{ $F_1$ Score for Cancer Class}} \\ \hline 
{Architecture\textbackslash Model} & {Model1} & {Model2} & {Model3} & {Model4} & {Model5} & {Model6} & {Model7} & {Majority Voting} \\ \hline
{SD-Net} & 0.877 & 0.908 & 0.904 & 0.872 & 0.875 & {0.885} & 0.891 & 0.8982 \\ 
{SDCT-Net} & {0.900} & {0.926} & {0.921} & {0.926} & {0.900} & 0.855 & {0.912} & \textbf{0.928} \\ \hline 
\multicolumn{9}{c}{{Weighted $F_1$ ($WF_1$) Score}} \\ \hline 
{Architecture\textbackslash Model} & {Model1} & {Model2} & {Model3} & {Model4} & {Model5} & {Model6} & {Model7} & {Majority Voting} \\ \hline
{SD-Net} & 0.849 & 0.883 & 0.879 & 0.840 & 0.848 & {0.859} & 0.868 & 0.873 \\ 
{SDCT-Net} & {0.873} & {0.904} & {0.894} & {0.904} & {0.875} & 0.829 & {0.885} & \textbf{0.908} \\ \hline 
\multicolumn{9}{c}{{Balanced Accuracy (\textit{BAC})}} \\ \hline 
{Architecture\textbackslash Model} & {Model1} & {Model2} & {Model3} & {Model4} & {Model5} & {Model6} & {Model7} & {Majority Voting} \\ \hline
{SD-Net} & 0.863 & 0.886 & {0.883} & 0.845 & 0.862 & {0.874} & {0.889} & 0.885 \\ 
{SDCT-Net} & {0.875} & {0.903} & {0.882} & {0.905} & {0.887} & 0.856 & {0.879} & \textbf{0.911} \\ \hline
\end{tabular}}
\end{small}
\end{table}

\begin{figure*}[!ht]
\centerline{
\begin{subfigure}[b]{0.43\textwidth}
\includegraphics[scale=0.37]{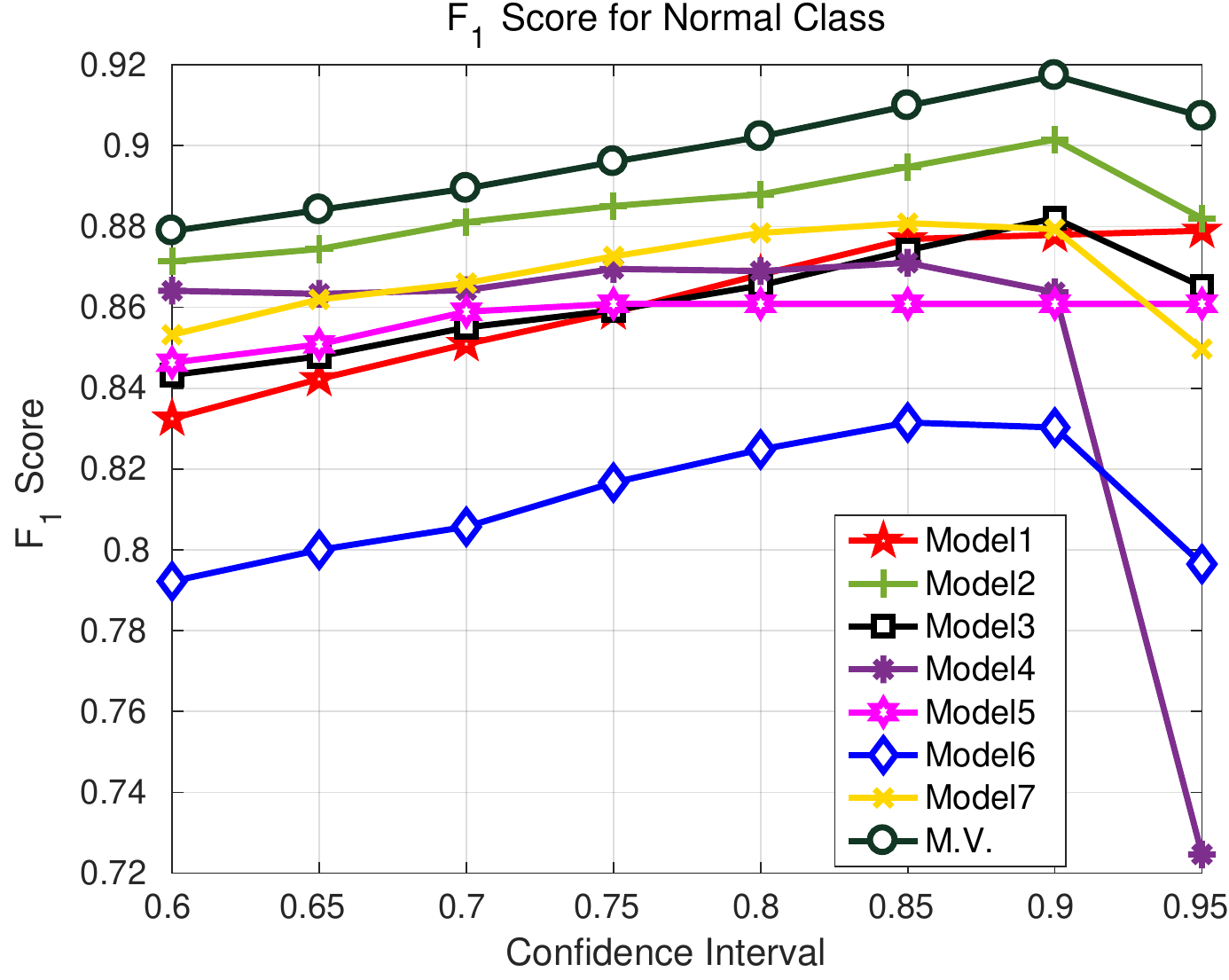}
\captionsetup{justification=centering}
\end{subfigure}
\begin{subfigure}[b]{0.43\textwidth}
\includegraphics[scale=0.38]{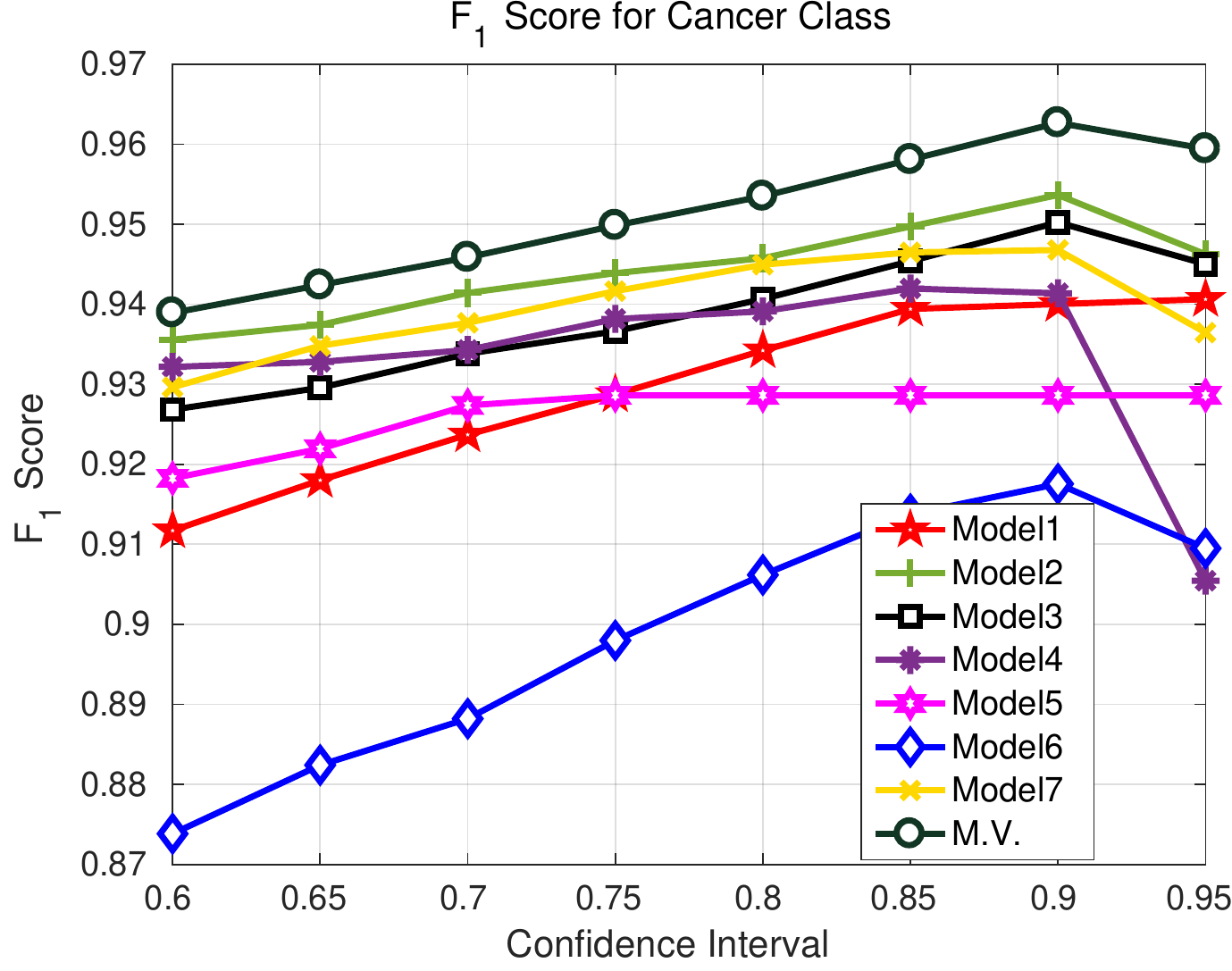}
\captionsetup{justification=centering}
\end{subfigure}
\begin{subfigure}[b]{0.43\textwidth}
\includegraphics[scale=0.38]{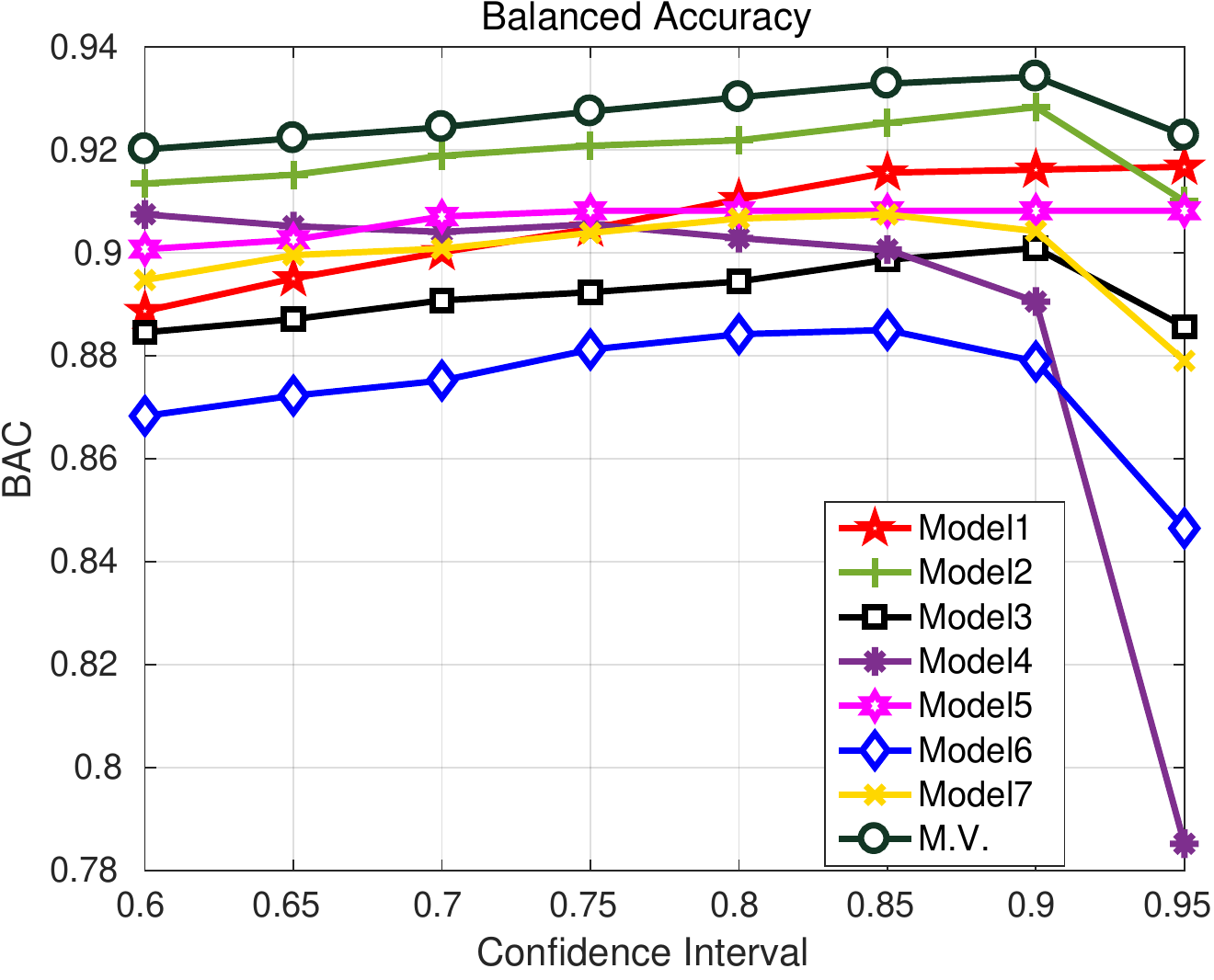}
\captionsetup{justification=centering}
\end{subfigure}}
\captionsetup{justification=centering}
\caption{Performance of SDCT-AuxNet$^{\theta}$ for different values of confidence probability ($\theta$): (a) $F_1$ score of normal class, (b) $F_1$ score of cancer class and (c) Balanced Accuracy. Model1 represents model trained with fold1 as the validation set and remaining folds as training set. Similar notation applies to Model2 to Model7. M.V is the score with majority voting on Model2 to Model7. Performance improves consistently with the increasing value of $\theta$ until $\theta=0.9$, after which a decline in the performance is noted. Results are presented on the test set of 2586 images.}
\label{effect of theta on SDCT-AuxNet}
\end{figure*}

\begin{table*}[!ht]
\vspace{1.5em}
\centering
\captionsetup{justification=centering}
\caption{Weighted $F_1$ score of SDCT-AuxNet$^{\theta}$ for different values of confidence probability ($\theta$). Model1 represents the model trained with fold1 as the validation set and remaining folds as training set. Similar notation applies to Model2 to Model7. The weighted $F_1$ ($WF_1$) score improves with increasing $\theta$ until $\theta=0.9$, beyond which the performance starts declining. Results are presented on the test set of 2586 images. Best results are indicated in bold case.}
\vspace{-1em}
\label{effect of theta on wf1 in SDCT-AuxNet}
\begin{small}
\centerline{
\begin{tabular}{ccccccccc}
\hline
\multicolumn{9}{c}{Weighted $F_1$ ($WF_1$) Score} \\ \hline
\begin{tabular}[c]{@{}c@{}}Confidence\\ Probability ($\theta$) \textbackslash Models\end{tabular} & Model1 & Model2 & Model3 & Model4 & Model5 & Model6 & Model7 & Majority Voting \\ \hline
0.6 & 0.886 & 0.915 & 0.900 & 0.910 & 0.895 & 0.848 & 0.905 & 0.920 \\ 
0.65 & 0.894 & 0.917 & 0.904 & 0.911 & 0.899 & 0.856 & 0.912 & 0.924 \\ 
0.7 & 0.900 & 0.922 & 0.909 & 0.912 & 0.906 & 0.862 & 0.915 & 0.928 \\ 
0.75 & 0.906 & 0.925 & 0.912 & 0.916 & 0.907 & 0.872 & 0.920 & 0.933 \\ 
0.8 & 0.913 & 0.927 & 0.917 & 0.917 & 0.907 & 0.880 & 0.924 & 0.937 \\ 
0.85 & 0.919 & 0.932 & 0.923 & 0.919 & 0.907 & 0.888 & 0.926 & 0.943 \\ 
0.9 & {0.920} & {0.937} & {0.928} & 0.917 & 0.907 & {0.890} & 0.925 & \textbf{0.948} \\ 
0.95 & 0.921 & 0.926 & 0.920 & 0.848 & 0.907 & 0.873 & 0.909 & 0.943 \\ \hline
\end{tabular}}
\end{small}
\end{table*}

\vspace{2em}
\subsection{Experiments with SDCT-AuxNet$^{\theta}$}
In this section, we analyze the performance of SDCT-AuxNet$^{\theta}$ architecture according to the testing strategy as discussed in Section \ref{section for testing strategy }. Support Vector Machine (SVM) with radial basis function (RBF) kernel is used as the auxiliary classifier. We analyze  SDCT-AuxNet$^{\theta}$ using different experiments: 1) by studying the impact of confidence probability $\theta$ on the performance of SDCT-AuxNet$^{\theta}$, 2) by studying the performance comparison of SDCT-AuxNet$^{\theta}$, SDCT-Net$^{0}$, and SDCT-Net$^{1}$, 3) by studying the subject-level performance of SDCT-AuxNet$^{\theta}$. 

\subsubsection{Impact of Confidence Probability } We experiment with different values of the confidence probability ($\theta$) to analyze its effect on the performance of the model. Results are summarized in Fig.~\ref{effect of theta on SDCT-AuxNet} and Table~\ref{effect of theta on wf1 in SDCT-AuxNet}. Fig.~\ref{effect of theta on SDCT-AuxNet} shows the effect of confidence probability on individual class $F_1$ score and balanced accuracy. It is observed that for the most of the models, performance improves with an increase of $\theta$ until $\theta=0.9$ after which a decline in performance is observed. The final prediction, i.e., prediction with majority voting also improves until $\theta=0.9$ and decreases thereafter. A similar pattern is observed for $BAC$, results for which are also summarized in Table~\ref{effect of theta on wf1 in SDCT-AuxNet}.

\subsubsection{Comparison of SDCT-AuxNet$^{\theta}$, SDCT-Net$^0$ and SDCT-Net$^1$}
We have compared the performance of SDCT-AuxNet$^{\theta}$ for $\theta=0.90$, $\theta=1$, and $\theta=0$ in Table~\ref{comparison of SDCT-Net and SDCT-AuxNet}. SDCT-AuxNet$^0$ and SDCT-AuxNet$^1$ correspond to the results with $\theta=0$ and $\theta=1$, respectively, in SDCT-AuxNet$^{\theta}$. This is to note that SDCT-AuxNet$^0$ is SDCT-AuxNet$^{\theta}$ without auxiliary classifier and hence, SDCT-AuxNet$^0$ is same as SDCT-Net. Similarly, SDCT-AuxNet$^1$ is SDCT-AuxNet$^{\theta}$ with only auxiliary classifier. Table~\ref{comparison of SDCT-Net and SDCT-AuxNet} shows the performance of SDCT-AuxNet$^{\theta}$ that exploits two different classifiers (SDCT-AuxNet$^0$ and SDCT-AuxNet$^1$ with different architectures) in a coupled fashion.
From the performance of individual models (Model1-Model7) of all these architectures, it can be inferred that SDCT-AuxNet$^{\theta}$ has achieved a significant gain over SDCT-AuxNet$^{0}$ and SDCT-AuxNet$^{1}$ in terms of all three performance metrics. Although majority voting is able to bridge the performance difference between SDCT-AuxNet$^{\theta}$ and SDCT-AuxNet$^{1}$, the former is leading the later for  most of the models (Model1-Model7), while the performance coincides in some cases. Compared to SDCT-AuxNet$^{0}$ (SDCT-Net), SDCT-AuxNet$^{\theta}$ provides a gain of $5.5\%$ for $F_1$ score of normal class, $3.5\%$ for $F_1$ score of cancer class, $4.0\%$ for weighted $F_1$ score, and $2.3\%$ for balanced accuracy. The highest gain is observed for $F_1$ score of normal class, which is $0.917$ with SDCT-AuxNet$^{\theta}$ as compared to $0.862$ with SDCT-AuxNet$^{0}$. To summarize, SDCT-AuxNet$^{\theta}$ ensembles the performance of SDCT-AuxNet$^{0}$ and SDCT-AuxNet$^{1}$ that results in significant performance improvement.

\begin{table*}[!t]
\begin{small}
\captionsetup{justification=centering}
\caption{Performance comparison of SDCT-AuxNet${^0}$, SDCT-AuxNet${^1}$, and SDCT-AuxNet${^\theta}$ for $\theta=0.9$. SDCT-Net is is a special case of SDCT-AuxNet${^\theta}$ with $\theta=0$ and is represented as SDCT-AuxNet$^{0}$. Similarly, SDCT-AuxNet${^1}$ is a spacial case of  SDCT-AuxNet${^\theta}$ with only the auxiliary classifier branch. Model1 represents the model trained with fold1 as the validation set and remaining folds as training set. Similar notation applies to Model2 to Model7. Best results are highlighted in black and equal performance is highlighted in blue color. SDCT-AuxNet${^\theta}$
performed better than SDCT-AuxNet${^0}$ or SDCT-AuxNet${^1}$. Results are presented on the test set of 2586 images.}
\vspace{-1em}
\label{comparison of SDCT-Net and SDCT-AuxNet}
\centerline{
\begin{tabular}{ccccccccc}
\hline
\multicolumn{9}{c}{$F_1$ Score for Normal Class} \\ \hline
Architecture\textbackslash Model & Model1 & Model2 & Model3 & Model4 & Model5 & Model6 & Model7 & Majority Voting \\ \hline
SDCT-AuxNet${^0}$ &  0.813 & 0.855 & 0.836 & 0.856 & 0.822 & 0.774 & 0.825 & 0.862 \\
SDCT-AuxNet${^1}$ & \textcolor{blue}{0.878} & 0.882& 0.865 & \textbf{0.868} & {0.860} & 0.794 & 0.849 & 0.915 \\
SDCT-AuxNet${^\theta}$ & \textcolor{blue}{0.878} & \textbf{0.902} & \textbf{0.882} & {0.864} & \textbf{0.861} & \textbf{0.830} & \textbf{0.879} & \textbf{0.917} \\ \hline
\multicolumn{9}{c}{$F_1$ Score for Cancer Class} \\ \hline
Architecture\textbackslash Model & Model1 & Model2 & Model3 & Model4 & Model5 & Model6 & Model7 & Majority Voting \\ \hline
SDCT-AuxNet${^0}$ & 0.900 & 0.926 & 0.921 & 0.926 & 0.900 & 0.855 & 0.912  & 0.928 \\
SDCT-AuxNet${^1}$ & \textcolor{blue}{0.940} & 0.946 & 0.945 & 0.936 & 0.928 & 0.908 & 0.936 & 0.961 \\
SDCT-AuxNet${^\theta}$ & \textcolor{blue}{0.940} & \textbf{0.954} & \textbf{0.950} & \textbf{0.941} & \textbf{0.929} & \textbf{0.918} & \textbf{0.947} & \textbf{0.963} \\ \hline
\multicolumn{9}{c}{Weighted $F_1$ ($WF_1$) Score} \\ \hline
Architecture\textbackslash Model & Model1 & Model2 & Model3 & Model4 & Model5 & Model6 & Model7 & Majority Voting \\ \hline
SDCT-AuxNet${^0}$ &  0.873 &  0.904 & 0.894 &  0.904 & 0.875 &  0.829 & 0.885 & 0.908 \\
SDCT-AuxNet${^1}$ & \textcolor{blue}{0.920} & 0.925 & 0.919 & 0.915& \textcolor{blue}{0.907} & 0.872 & 0.908 & 0.947 \\
SDCT-AuxNet${^\theta}$ & \textcolor{blue}{0.920} & \textbf{0.937} & \textbf{0.928} & \textbf{0.917} & \textcolor{blue}{0.907} & \textbf{0.89} & \textbf{0.925} & \textbf{0.948} \\ \hline
\multicolumn{9}{c}{Balanced Accuracy ($BAC$)} \\ \hline
Architecture\textbackslash Model & Model1 & Model2 & Model3 & Model4 & Model5 & Model6 & Model7 & Majority Voting \\ \hline
SDCT-AuxNet${^0}$ & {0.875} & { 0.903} & { 0.882} & {0.905} & { 0.887} & { 0.856} & {0.879} & {0.911} \\
SDCT-AuxNet${^1}$ & \textcolor{blue}{0.916} & {0.910} & {0.885} & \textbf{0.906} & \textcolor{blue}{0.908} & { 0.845} & {0.879} & {0.932} \\
SDCT-AuxNet${^\theta}$  & \textcolor{blue}{0.916} & \textbf{0.928} & \textbf{0.901} & {0.891} & \textcolor{blue}{0.908} & \textbf{0.879} & \textbf{0.904} & \textbf{0.934} \\ \hline
\end{tabular}}
\end{small}
\end{table*} 
\subsubsection{Subject-Level Performance of SDCT-AuxNet$^{\theta}$}
\begin{figure*}[!t]
\centerline{
\begin{subfigure}[b]{0.75\textwidth}
\includegraphics[scale=.8]{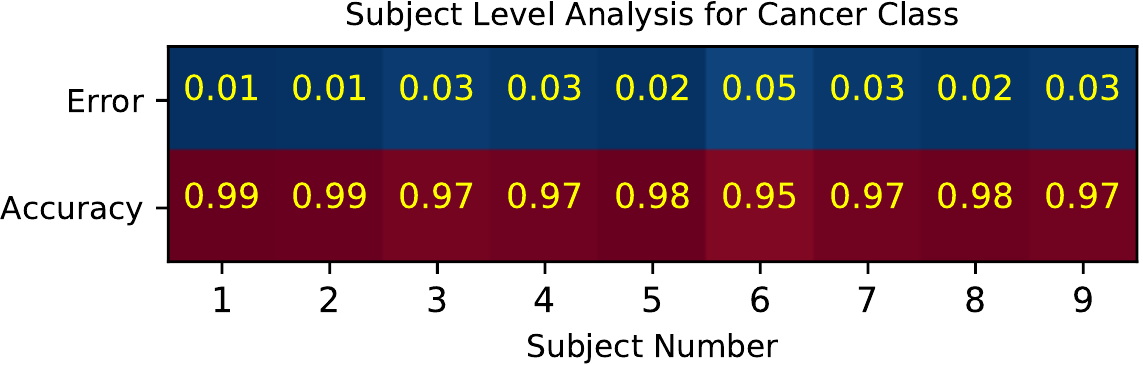}
\captionsetup{justification=centering}
\end{subfigure}
\begin{subfigure}[b]{0.75\textwidth}
\includegraphics[scale=.79]{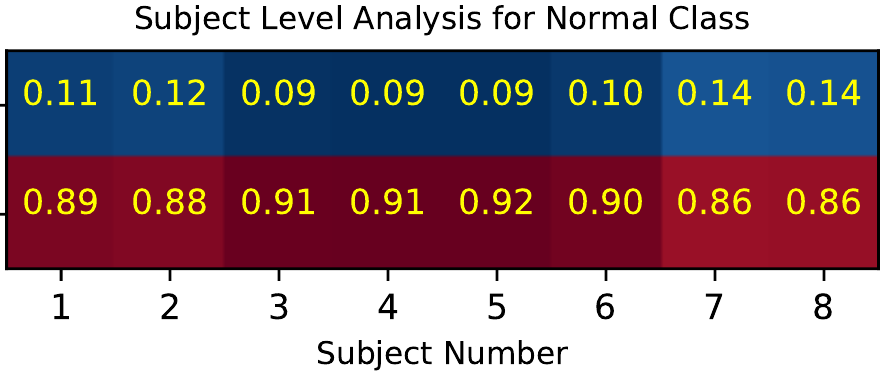}
\captionsetup{justification=centering}
\end{subfigure}}
\captionsetup{justification=centering}
\caption{Subject-level performance of SDCT-AuxNet$^{\theta}$ is observed to be consistent across all the subjects indicating that SDCT-AuxNet$^{\theta}$ is robust to subject-level variations.}
\label{subject level analysis}
\end{figure*}

We have also evaluated the performance of SDCT-AuxNet$^{\theta}$ for individual subjects in terms of accuracy. The subject-level accuracy and error rate for both the classes is depicted in Fig.~\ref{subject level analysis}. SDCT-AuxNet$^{\theta}$ performs consistently good for all the subjects. These results are encouraging since it shows that the model is robust to subject-level variability. 

\subsection{Comparison with the state-of-the-art Methods}
This C-NMC 2019 dataset has recently been used by \cite{JoansCNMC02} and \cite{LiuCNMC01}. \cite{JoansCNMC02} have used training set to fine-tune a pretrained SE-ResNeXt50 architecture. \cite{LiuCNMC01} have divided the larger class samples into two subsets, which are then combined with other class samples to train two different models initialized with pretrained Inception ResNets weights. These two models are then fine-tuned jointly on whole training dataset to obtain the final predictions. These methods are able to achieve a weighted $F_1$ score of $87.6\%$ \citep{LiuCNMC01}, and $88.9\%$ \citep{JoansCNMC02} as compared to $94.8\%$ of the proposed method. 
Apart from these two approaches, other participants have used methodologies based on VGGNet, AlexNet, InceptionV3, DenseNet and MobileNet which have performed below the challenge winner's best performance of 91\%.

We also compared our results with the top entry at the final leaderboard of the challenge \citep{ref50}. The leaderboard shows the performance on final test set in terms of weighted $F_1$ score. It is observed from Table~\ref{comaprison with cnmc} that SDCT-AuxNet$^{\theta}$ is performing better than top entry method  with a significant margin of $3.8\%$. 
\begin{table}[!h]
\centering
\captionsetup{justification=centering}
\caption{Comparison of SDCT-Net and SDCT-AuxNet$^{\theta}$ with the top entry of C-NMC 2019 Challenge. Best results are depicted in bold case.}
\begin{small}
\vspace{-1em}
\label{comaprison with cnmc}
\centerline{
\begin{tabular}{cccc }
\hline
Metric\textbackslash Architecture & C-NMC & SDCT-Net & SDCT-AuxNet$^{\theta}$ \\ 
\hline
Weighted $F_1$ ($WF_1$) & 0.910  & 0.908 & \textbf{0.948}\\ \hline
\end{tabular}}
\end{small}
\end{table}


\section{Discussion}
\begin{figure*}[!ht]
\centerline{
\includegraphics[scale=0.9]{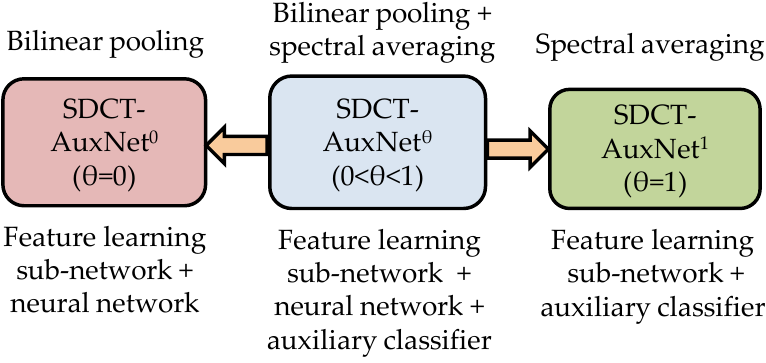}}
\captionsetup{justification=centering}
\vspace{-1em}
\caption{SDCT-AuxNet$^{\theta}$ as a generic case utilizing two different classifiers in decision making. SDCT-Net is same as SDCT-AuxNet$^{\theta}$ for $\theta=0$ and utilizes only neural network. SDCT-AuxNet$^{1}$ is SDCT-AuxNet$^{\theta}$ with $\theta=1$ and use only the auxiliary classifier. SDCT-AuxNet$^{\theta}$ swings between SDCT-AuxNet$^{0}$ (SDCT-Net) and SDCT-AuxNet$^{1}$ depending upon the value of confidence probability $\theta$.}
\label{sdctauxnet}
\end{figure*}

In this section, we have analyzed SDCT-AuxNet$^{\theta}$ from different perspectives. First, we analyze SDCT-AuxNet$^{\theta}$ as a generic architecture that can be modified to two different architectures through the setting of confidence probability ($\theta$). As discussed in above sections, confidence probability ($\theta$) that lies in the range $[0,1]$, plays a crucial role during the testing phase of SDCT-AuxNet$^{\theta}$. 
By setting the value of $\theta$, we are deciding the contribution of two different classifiers in decision making. However, with two extreme values of $\theta$, i.e. $\theta=0$  or $\theta=1$, we can also choose only one classifier. This scenario is depicted in Fig.~\ref{sdctauxnet}. By setting $\theta=0$, SDCT-AuxNet$^{\theta}$ is reduced to SDCT-AuxNet$^{0}$ (SDCT-Net), which is the feature learning sub-network along with neural network as the classifier that utilizes bilinear pooling. For $\theta=1$, SDCT-AuxNet$^{\theta}$ becomes  SDCT-AuxNet$^{1}$, which utilizes feature learning sub-network along with spectral averaging and auxiliary classifier. Between these two extremes, SDCT-AuxNet$^{\theta}$ exploits bilinear pooling with neural network  and spectral averaging with auxiliary classifier. This methodology is an advantage of SDCT-AuxNet$^{\theta}$ because it provides flexibility to decide the contribution of the classifiers by simply setting the value of $\theta$. This flexibility is also the reason for the better performance of SDCT-AuxNet$^{\theta}$  over SDCT-AuxNet$^{0}$ (SDCT-Net) or SDCT-AuxNet$^{1}$. In fact, the coupling of SDCT-AuxNet$^{0}$ (SDCT-Net) and SDCT-AuxNet$^{1}$ in SDCT-AuxNet$^{\theta}$ can also be interpreted as an ensemble of two classifiers, with each classifier utilizing a different set of features. Results of Table~\ref{comparison of SDCT-Net and SDCT-AuxNet} validates our hypothesis that an appropriate coupling of two different classifiers (working on different sets of features) may lead to better performance as compared to an individual classifier utilizing a single set of features. 

\begin{table}[!t]
\vspace{1em}
\centering
\captionsetup{justification=centering}
\caption{The number of samples predicted by different classifiers at $\theta=0.9$. Contribution  of each classifier is approximately the same in majority of the cases, indicating significant role of both the classifiers in decision making.}
\vspace{-1em}
\label{contribution of classifiers}
\small
\begin{tabular}{ccccclll} \hline
\begin{tabular}[c]{@{}c@{}}Classifier/\\ Models\end{tabular} & Model1 & Model2 & Model3 & Model4 & Model5 & Model6 & Model7 \\ \hline
\begin{tabular}[c]{@{}c@{}}Neural \\ Network\end{tabular} & 1634 & 1396 & 428 & 1885 & 1199 & 1544 & 1056 \\ \hline
\begin{tabular}[c]{@{}c@{}}Auxiliary \\ Classifier\end{tabular} & 952 & 1190 & 2158 & 701 & 1387 & 1042 & 1530 \\ \hline
Total Samples & \multicolumn{7}{c}{2586} \\ \hline
\end{tabular}
\end{table}

We have also analyzed the contribution of each classifier in the final prediction (Table~\ref{contribution of classifiers}). It is observed that both the classifiers contribute approximately equally in most of the cases. These results prove the hypothesis that both the classifiers are involved in decision making, and the final outcome is not due to only one of the classifiers.

\begin{figure*}[!ht]
\centerline{
\begin{subfigure}[b]{0.31\textwidth}
\includegraphics[scale=0.5]{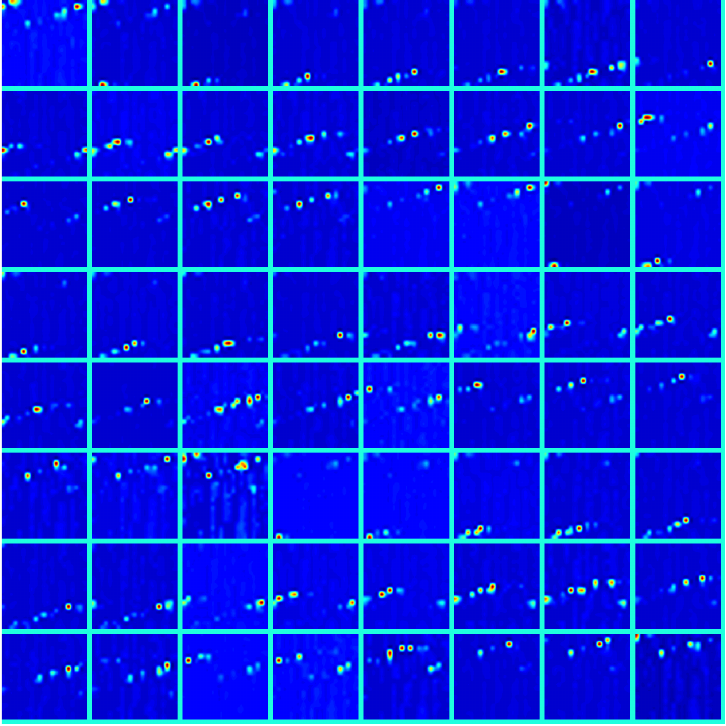}
\captionsetup{justification=centering}
\caption{Features of Cancer Class }
\end{subfigure}
\begin{subfigure}[b]{0.31\textwidth}
\includegraphics[scale=0.5]{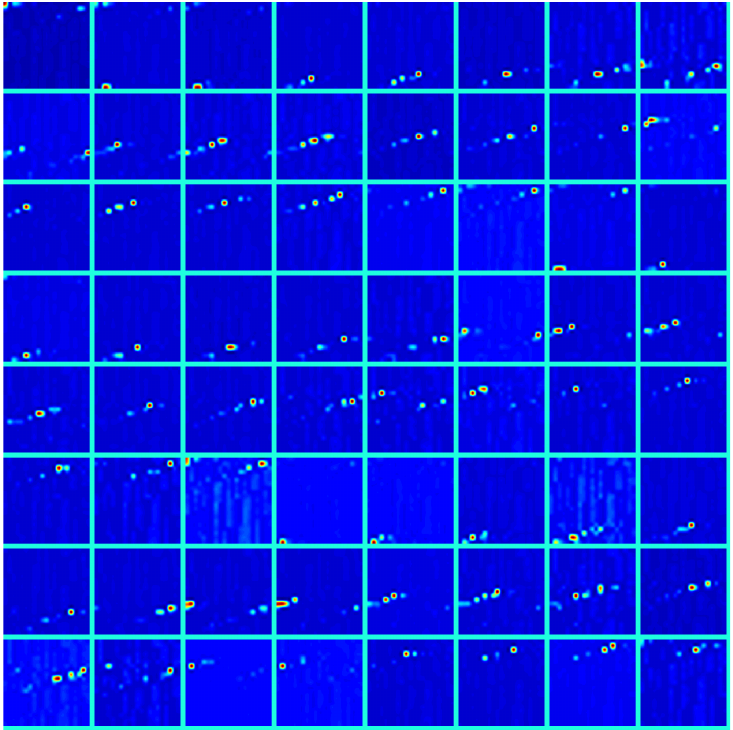}
\captionsetup{justification=centering}
\caption{Features of normal Class }
\end{subfigure}
\begin{subfigure}[b]{0.31\textwidth}
\includegraphics[scale=0.553]{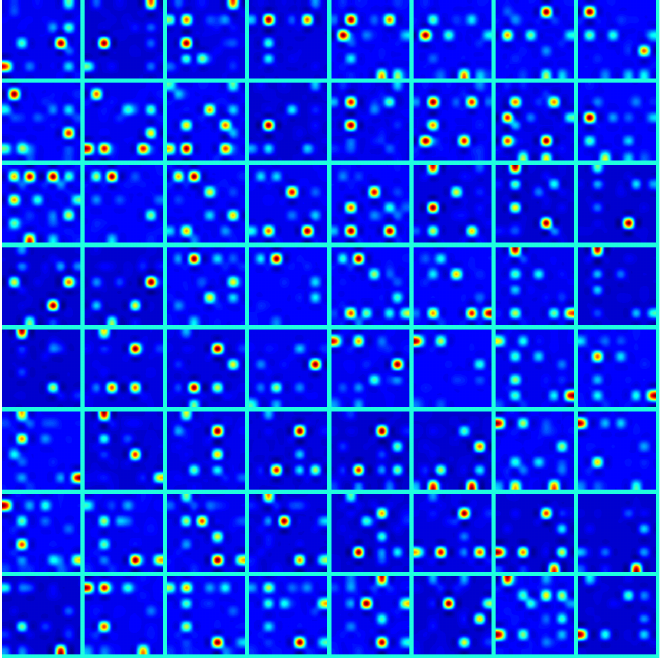}
\captionsetup{justification=centering}
\caption{Features of Cancer Class }
\end{subfigure}
\begin{subfigure}[b]{0.31\textwidth}
\includegraphics[scale=0.545]{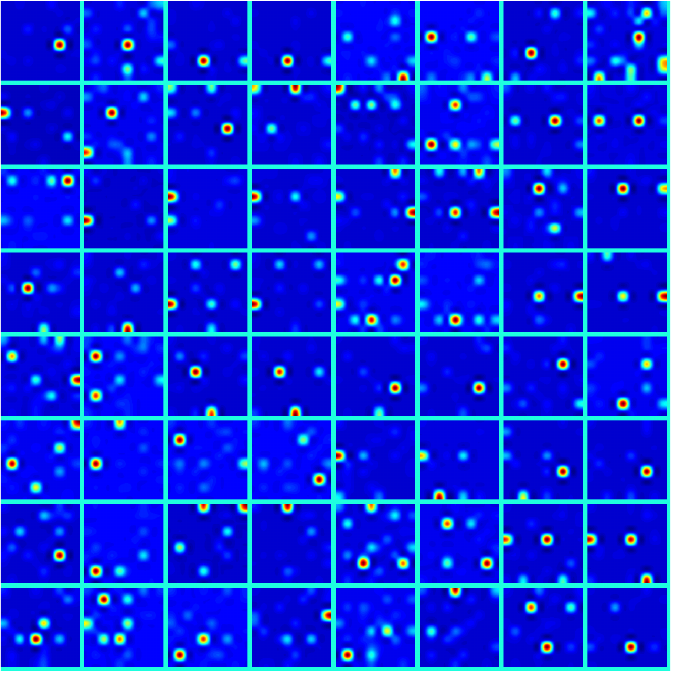}
\captionsetup{justification=centering}
\caption{Features of normal Class }
\end{subfigure}}
\captionsetup{justification=centering}
\caption{Feature maps of second last (C5) and last convolutional  layer (C6) for cancer class (a,c) and normal class (b,d). The features for normal class are more sparse as compared to the features of cancer class. This difference in sparsity helps to improve the performance of the classifier.  }
\label{last layer features}
\end{figure*}

Next, we visualize the features of the second last convolutional layer and the final convolutional layer of SDCT-AuxNet$^{\theta}$  in order to understand if something different is learned by these layers. In Fig.~\ref{dct output}, we visualized the sparsity of the normal cell images and cancer cell images in the DCT domain. Normal cell images are observed to be sparsified more as compared to cancer cell images. We claimed that the difference in sparsity would help CNN to distinguish the two classes. This effect is propagated to the deeper layers as visualized in Fig.~\ref{last layer features}, where it is again observed that the features for the normal class are more sparse as compared to the cancer class. This observation implies that different sparsification induced by the DCT-layer may be one of the essential characteristics of class discrimination and hence, SDCT-AuxNet$^{\theta}$ could separate the two classes based on sparsity, even if the images of both classes are visually similar. This property is crucial for images in medical applications such as ALL classification because images are not visually differentiable. Hence, features like sparsity difference may help to enhance the performance of the classifier.

\begin{figure*}[!ht]
\vspace{1em}
\centerline{
\begin{subfigure}[b]{0.5\textwidth}
\centering
\includegraphics[scale=0.4]{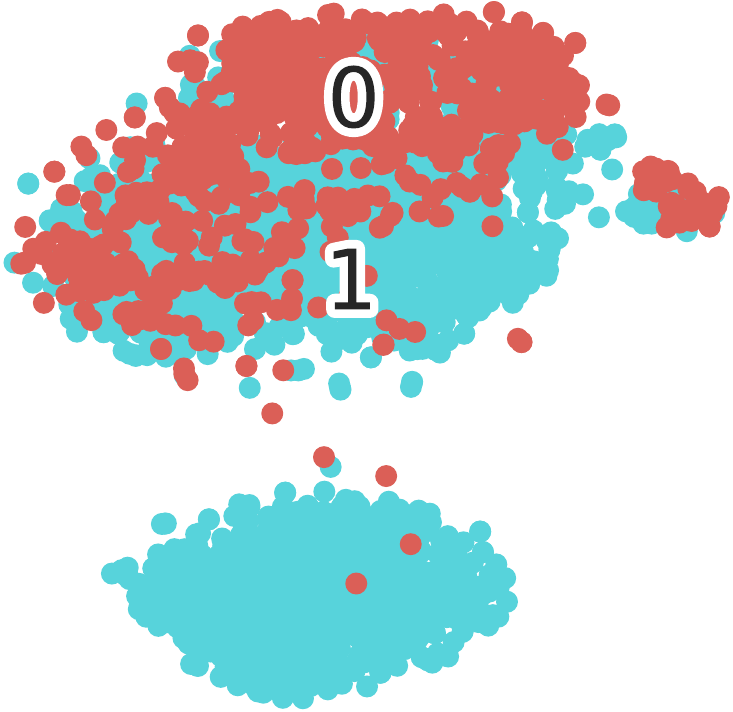}
\captionsetup{justification=centering}
\caption{First Convolutional Layer (C1)}
\end{subfigure}
\begin{subfigure}[b]{0.5\textwidth}
\centering
\includegraphics[scale=0.4]{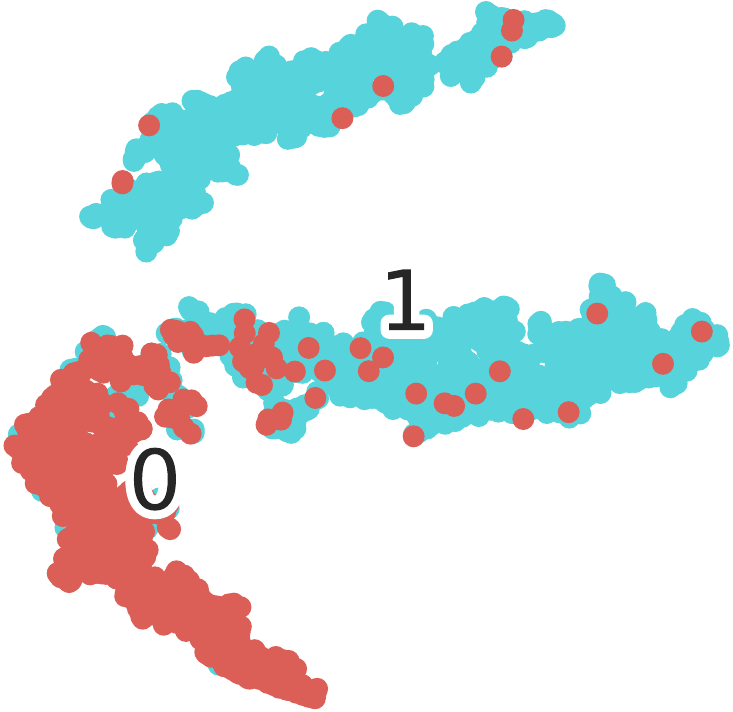}
\captionsetup{justification=centering}
\caption{Sixth Convolutional Layer (C6)}
\end{subfigure}}
\captionsetup{justification=centering}
\caption{Class discrimination at first (C1) and last convolutional layers (C6). The samples are separated as we progress deeper into the network. At the final layer, majority of the samples can be easily separated even with a linear boundary.}
\label{layers output}
\end{figure*}

We have also visualized class discrimination at different layers in Fig.~\ref{layers output}. At the first convolutional layer, majority of the samples of the two classes are overlapping. However, as we go deeper into the network, the two classes start to diverge. Finally, at the last convolutional layer, the two classes can be linearly separated. This points to an intuitive understanding that difference in sparsity in SDCT-AuxNet$^{\theta}$ may be one of the significant causes of  separability of the two classes. Although some samples still lie on the opposite side of the boundary, majority of the samples can be classified correctly. These observations establish the validity of DCT-layer in ALL classification. 
 
Finally, samples in the medical domain are collected from different subjects, which may induce subject-level variations. Robustness to subject-level variability is a crucial factor for the practical implementation of the classifier. To mimic this effect, we trained and tested the architectures on folds separated at subject level. Results summarized in Fig.~\ref{subject level analysis} prove the robustness of SDCT-AuxNet$^{\theta}$ to subject-level variability. However, error rate for normal class is high as compared to cancer class. This results imply that classifier has relatively lower specificity as compared to recall. Hence, classifier yields excellent performance in predicting the disease in unhealthy subjects, but relatively poor performance in discarding the disease in healthy subjects.

\section{Conclusion and Future Work}

We proposed a novel framework, SDCT-AuxNet$^{\theta}$, for the challenging problem of ALL classification. The framework utilizes two different classifiers operating on two different set of features. SDCT-AuxNet$^{\theta}$  use sparsity in DCT domain to learn class discriminative features. In ALL classification, it is important to pool the data at subject-level due to inter-subject variability in cell images. We have used this approach to make SDCT-AuxNet$^{\theta}$ robust to subject-level variations. However, there is some performance margin between the performance for normal and cancer class. We will cover this issue in our future work. We also plan to design a multi task framework for cancer diagnosis. 

\section{Acknowledgements}
\vspace{-0.5em}
Shiv Gehlot would like to thank University Grant Commission (UGC),
Govt. of India for the UGC-Junior Research Fellowship. We also acknowledge the Infosys Center for Artificial Intelligence, IIIT-Delhi for our research work.





\section*{References}

\bibliography{c_nmc_preprint}

\end{document}